\documentclass[10pt,twocolumn,letterpaper]{article}

\usepackage{cvpr}
\usepackage{times}
\usepackage{epsfig}
\usepackage{graphicx}
\usepackage{amsmath}
\usepackage{amssymb}

\usepackage{url}
\usepackage{epsfig}
\usepackage{mathrsfs}
\usepackage{multirow}
\usepackage{amsmath,bm}
\usepackage{tabu}
\usepackage{rotating}
\usepackage{stfloats}
\usepackage{float}

\usepackage[breaklinks=true,bookmarks=false]{hyperref}

\cvprfinalcopy 

\def\cvprPaperID{****} 

\begin{document}

\title{Instance-aware Image and Sentence Matching with Selective Multimodal LSTM}

\author{Yan Huang$^1$ \hspace{7mm} Wei Wang$^1$  \hspace{7mm} Liang Wang$^{1,2}$\\
$^1$Center for Research on Intelligent Perception and Computing\\
National Laboratory of Pattern Recognition\\
$^2$Center for Excellence in Brain Science and Intelligence Technology\\
Institute of Automation, Chinese Academy of Sciences\\
{\tt\small \{yhuang, wangwei, wangliang\}@nlpr.ia.ac.cn}
}


\makeatletter
\g@addto@macro\@maketitle{
  \vspace{-10mm}
  \begin{figure}[H]
  \setlength{\linewidth}{\textwidth}
  \setlength{\hsize}{\textwidth}
  \centering
  \includegraphics[scale=0.18]{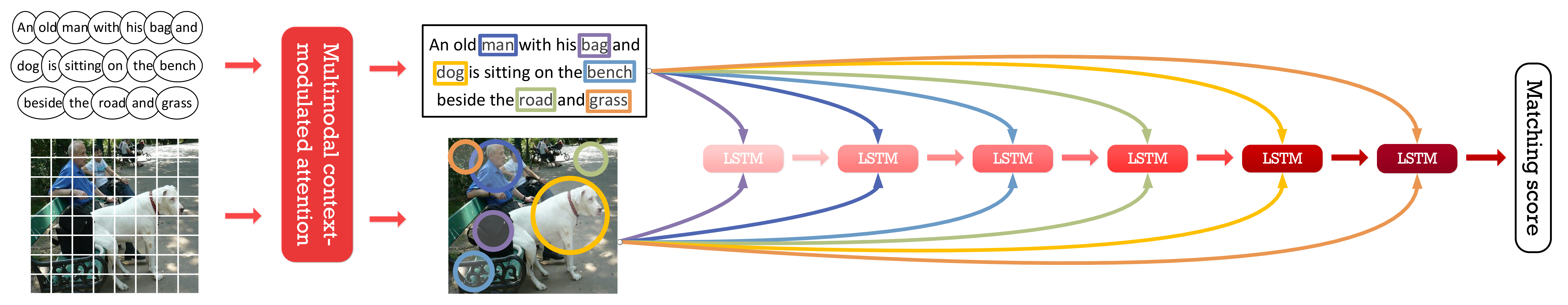}
  \caption{The proposed selective multimodal Long Short-Term Memory network (sm-LSTM) (best
  viewed in colors).}
  \label{fig:sm-LSTM}
  \end{figure}
}
\makeatother
\maketitle

\begin{abstract}
Effective image and sentence matching depends on how to well measure
their global visual-semantic similarity.
Based on the observation that
such a global similarity arises from a complex aggregation of
multiple local similarities between
pairwise instances of image (objects) and sentence (words),
we propose a selective multimodal Long Short-Term Memory network (sm-LSTM)
for instance-aware image and sentence matching.
The sm-LSTM includes a multimodal context-modulated attention scheme at each timestep
that can selectively attend to a pair of instances of image and sentence,
by predicting pairwise instance-aware saliency maps for image and sentence.
For selected pairwise instances, their representations are obtained based on
the predicted saliency maps, and then compared to measure their local similarity.
By similarly measuring multiple local similarities within a few timesteps,
the sm-LSTM sequentially aggregates them with hidden states
to obtain a final matching score as the desired global similarity.
Extensive experiments show that
our model can well match image and sentence with complex content,
and achieve the state-of-the-art results on two public benchmark datasets.
\end{abstract}

\section{Introduction}


Matching image and sentence plays an important role in many applications,
e.g., finding sentences given an image query for image annotation and caption,
and retrieving images with a sentence query for image search.
The key challenge of such a cross-modal matching task is
how to accurately measure the image-sentence similarity.
Recently, various methods have been proposed for this problem,
which can be classified into two categories:
1) one-to-one matching and 2) many-to-many matching.

One-to-one matching methods usually extract
global representations for image and sentence,
and then associate them using either a structured objective \cite{frome2013devise,kiros2014unifying,vendrov2015order}
or a canonical correlation objective \cite{yan2015deep,klein2015associating}.
But they ignore the fact that
the global similarity commonly arises from a
complex aggregation of local similarities
between image-sentence instances (objects in an image and words in a sentence).
Accordingly, they fail to perform accurate instance-aware image and sentence matching.



Many-to-many matching methods \cite{karpathy2014deep,karpathy2014vsa,plummer2015flickr30k}
propose to compare many pairs of image-sentence instances,
and aggregate their local similarities.
However, it is not optimal to measure local similarities for all the possible
pairs of instances without any selection,
since only partial salient instance pairs describing the same semantic concept
can actually be associated and contribute to the global similarity.
Other redundant pairs are less useful which could
act as noise that degenerates the final performance.
In addition, it is not easy to obtain instances for either image or sentence,
so these methods usually have to explicitly employ additional object detectors \cite{de2006generating}
and dependency tree relations \cite{girshick2014rich}, or expensive human annotations.

To deal with these issues mentioned above,
we propose a sequential model, named selective multimodal Long Short-Term Memory
network (sm-LSTM),
that can recurrently select salient pairs of image-sentence instances,
and then measure and aggregate their local similarities within several timesteps.
As shown in Figure \ref{fig:sm-LSTM},
given a pair of image and sentence with complex content,
the sm-LSTM first extracts their instance candidates, i.e.,
words of the sentence and regions of the image.
Based on the extracted candidates, the model exploits a multimodal context-modulated attention
scheme at each timestep to selectively attend to a pair of desired image
and sentence instances (marked by circles and rectangles with the same color).
In particular, the attention scheme first predicts pairwise
instance-aware saliency maps for the image and sentence,
and then combines saliency-weighted representations of candidates to represent the attended pairwise instances.
Considering that each instance seldom occurs in isolation but
co-varies with other ones as well as the particular context,
the attention scheme uses multimodal global context as reference
information to guide instance selection.

Then, the local similarity of the attended pairwise instances can
be measured by comparing their obtained representations.
During multiple timesteps, the sm-LSTM exploits hidden states to
capture different local similarities of
selected pairwise image-sentence instances, and sequentially accumulates them
to predict the desired global similarity (i.e., the matching score) of image and sentence.
Our model jointly performs pairwise instance selection, local similarity learning and aggregation in a single framework,
which can be trained from scratch in an end-to-end manner
with a structured objective.
To demonstrate the effectiveness of the proposed sm-LSTM,
we perform experiments of image annotation and retrieval on two publicly available datasets,
and achieve the state-of-the-art results.


\section{Related Work}


\subsection{One-to-one Matching}

Frome \etal \cite{frome2013devise} propose a deep image-label embedding framework
that uses Convolutional Neural Networks (CNN) \cite{krizhevsky2012imagenet}
and Skip-Gram \cite{mikolov2013efficient} to extract representations for image and label, respectively,
and then associates them with a structured objective
in which the matched image-sentence pairs have small distances.
With a similar framework, Kiros \etal \cite{kiros2014unifying}
use Recurrent Neural Networks (RNN) \cite{hochreiter1997long} for sentence representation learning,
Vendrov \etal \cite{vendrov2015order} refine the objective
to preserve the partial order structure of visual-semantic hierarchy,
and Wang \etal \cite{wang2015learning} combine cross-view and within-view constraints
to learn structure-preserving embedding.
Yan \etal \cite{yan2015deep} associate representations of image and sentence
using deep canonical correlation analysis
where the matched image-sentence pairs have high correlation.
Using a similar objective, Klein \etal \cite{klein2015associating}
propose to use Fisher Vectors (FV) \cite{perronnin2007fisher}
to learn discriminative sentence representations,
and Lev \etal \cite{lev2015rnn} exploit RNN to encode FV for
further performance improvement.

\subsection{Many-to-many Matching}

Karpathy \etal \cite{karpathy2014vsa,karpathy2014deep} make the first attempt
to perform local similarity learning between
fragments of images and sentences with a structured objective.
Plummer \etal \cite{plummer2015flickr30k} collect region-to-phrase correspondences
for instance-level image and sentence matching.
But they indistinctively use all pairwise instances for similarity measurement,
which might not be optimal since there exist many matching-irrelevant instance pairs.
In addition, obtaining image and sentence instances is not trial,
since either additional object detectors or expensive human annotations
need to be used.
In contrast, our model can automatically select
salient pairwise image-sentence instances,
and sequentially aggregate their local similarities to obtain global similarity.

Other methods for image caption
\cite{mao2014explain,fang2015captions,donahue2015long,vinyals2015show,chen2015mind}
can be extended to deal with image-sentence matching,
by first generating the sentence given an image and then
comparing the generated sentence and groundtruth one word-by-word in a many-to-many manner.
But this kind of models are especially designed to
predict a grammar-completed sentence close to the groundtruth sentence,
rather than select salient pairwise sentence instances for similarity measurement.



\subsection{Deep Attention-based Models}

Our proposed model is related to some attention-based models.
Ba \etal \cite{ba2014multiple} present a recurrent attention model
that can attend to some label-relevant image regions of an image for multiple objects recognition.
Bahdanau \etal \cite{bahdanau2014neural} propose a neural machine translator which
can search for relevant parts of a source sentence to predict a target word.
Xu \etal \cite{xu2015show} develop an attention-based caption model
which can automatically learn to fix gazes on salient objects in an image
and generate the corresponding annotated words.
Different from these models, our sm-LSTM focuses on
joint multimodal instance selection and matching,
which uses a multimodal context-modulated attention scheme
to jointly predict instance-aware saliency maps for both image and sentence.



\begin{figure*}[t]
\begin{center}
\includegraphics[scale=0.37]{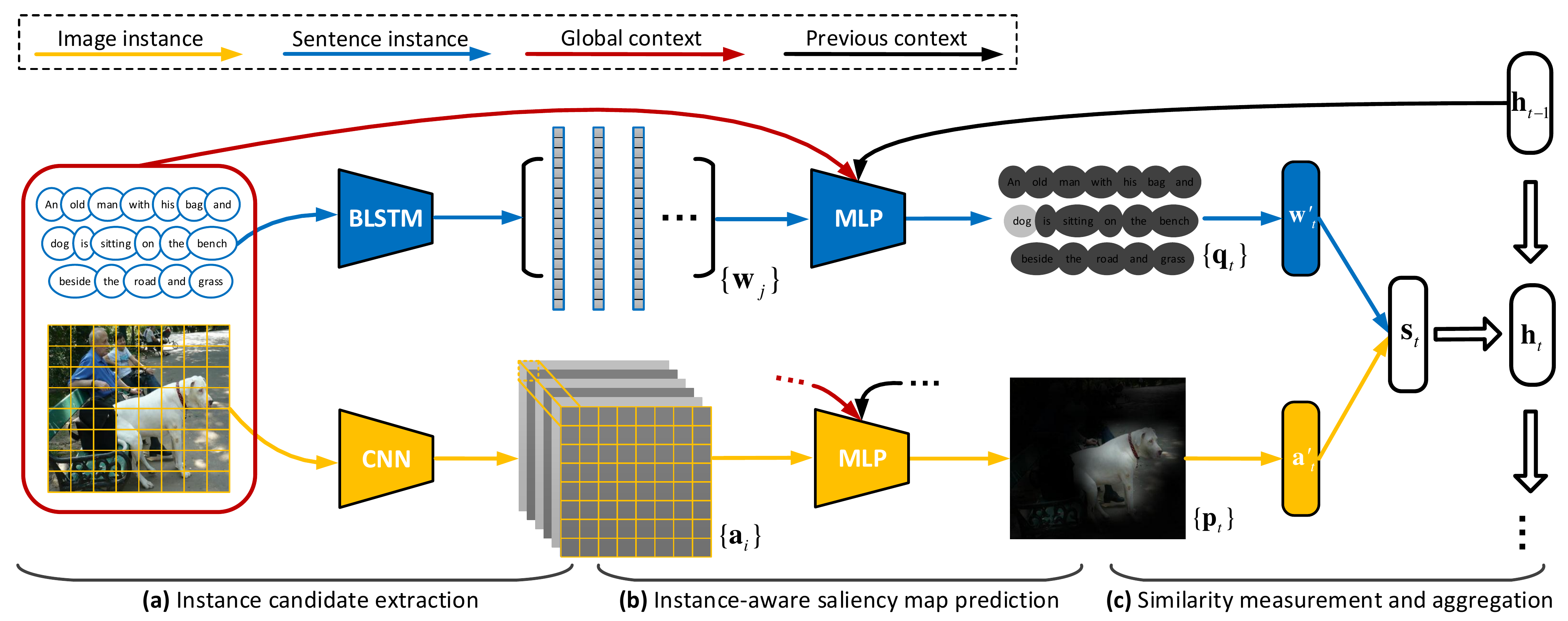}
\end{center}
\caption{Details of the proposed sm-LSTM, including (a) instance candidate extraction, (b) instance-aware saliency map prediction,
and (c) similarity measurement and aggregation (best viewed in colors).}
\label{fig:details}
\end{figure*}

\section{Selective Multimodal LSTM}
We will present the details of the proposed selective multimodal Long Short-Term Memory network (sm-LSTM)
from the following three aspects: (a) instance candidate extraction for both image and sentence,
(b) instance-aware saliency map prediction with a multimodal context-modulated attention scheme,
and (c) local similarity measurement and aggregation with a multimodal LSTM.


\subsection{Instance Candidate Extraction} \label{sect:candidates}

\textbf{Sentence Instance Candidates.}
For a sentence, its underlying instances mostly exist in word-level or phrase-level,
e.g., dog and man.
So we simply tokenlize and split the sentence into words,
and then obtain their representations by
sequentially processing them with a bidirectional LSTM (BLSTM) \cite{schuster1997bidirectional},
where two sequences of hidden states with different directions (forward and backward) are learnt.
We concatenate the vectors of two directional hidden states at the same timestep
as the representation for the corresponding input word.

\textbf{Image Instance Candidates.}
For an image, directly obtaining its instances is very difficult,
since the visual content is unorganized where the instances
could appear in any location with various scales.
To avoid the use of additional object detectors,
we evenly divide the image into regions as instance candidates
as shown in Figure \ref{fig:details} (a),
and represent them by extracting feature maps of the last convolutional layer in a CNN.
We concatenate feature values at the same location across different
feature maps as the feature vector for the corresponding convolved region.

\subsection{Instance-aware Saliency Map Prediction}



Apparently, neither the split words nor evenly divided
regions can precisely describe the desired sentence or image instances.
It is attributed to the fact that:
1) not all instance candidates are necessary since
both image and sentence consist of too much instance-irrelevant information,
and 2) the desired instances usually exist as a combination of
multiple candidates, e.g., the instance dog covers about 12 image regions.
Therefore, we have to evaluate the instance-aware saliency of each instance candidates,
with the aim to highlight those important and ignore those irrelevant.


To achieve this goal,
we propose a multimodal context-modulated attention scheme
to predict pairwise instance-aware saliency maps for image and sentence.
Different from \cite{xu2015show},
this attention scheme is designed for multimodal data rather than unimodal data,
especially for the multimodal matching task.
More importantly, we systematically study the importance of global context modulation
in the attentional procedure.
It results from an observation that
each instance of image or sentence seldom occurs in isolation but co-varies with other
ones as well as particular context.
In particular, previous work \cite{oliva2007role} has shown that the global image scene
enables humans to quickly guide their attention to regions of interest.
A recent study \cite{ghosh2016contextual} also demonstrates that the
global sentence topic capturing long-range context can greatly
facilitate inferring about the meaning of words.


As illustrated in Figure \ref{fig:details} (b),
we denote the previously obtained instance candidates of image and sentence
as $\left\{ {{{\textbf{a}}_i}}| {{\textbf{a}}_i} \in {\mathbb{R}^{F}}  \right\}_{i=1,\cdots,I}$
and $\left\{ {{{\textbf{w}}_j}}| {{\textbf{w}}_j} \in {\mathbb{R}^{G}}  \right\}_{j=1,\cdots,J}$, respectively.
${{\textbf{a}}_i}$ is the representation of the $i$-th divided region in the image
and $I$ is the total number of regions.
${{\textbf{w}}_j}$ describes the $j$-th split word in the sentence
and $J$ is the total number of words.
$F$ is the number of feature maps in the last convolutional layer of CNN
while $G$ is twice the dimension
of hidden states in the BLSTM.
We regard the output vector of the last fully-connected layer in the CNN
as the global context $\textbf{m} \in {\mathbb{R}^{D}}$ for the image,
and the hidden state at the last timestep in a sentence-based LSTM as
the global context $\textbf{n} \in {\mathbb{R}^{E}}$ for the sentence.
Based on these variables, we can perform
instance-aware saliency map prediction at the $t$-th timestep as follows:
\begin{equation} \label{eqn:e3}
\begin{aligned}
{p_{t,i}} = {{{e^{{{\hat p}_{t,i}}}}}}/{{\sum\nolimits_{i = 1}^I {{e^{{{\hat p}_{t,i}}}}} }}, {\kern 3pt}
{{\hat p}_{t,i}} = {f_{{p}}}(\textbf{m}, {\textbf{a}_i}, {{\textbf{h}}_{t - 1}}), \\
{q_{t,j}} = {{{e^{{{\hat q}_{t,j}}}}}}/{{\sum\nolimits_{j = 1}^J {{e^{{{\hat q}_{t,j}}}}} }}, {\kern 3pt}
{{\hat q}_{t,j}} = {f_{{q}}}(\textbf{n}, {\textbf{w}_j}, {{\textbf{h}}_{t - 1}})
\end{aligned}
\end{equation}
where ${p_{t,i}}$ and ${q_{t,j}}$ are saliency values indicating probabilities that the $i$-th image instance candidate
and the $j$-th sentence instance candidate
will be attended to at the $t$-th timestep, respectively.
$f_{{p}}(\cdot)$ and $f_{{q}}(\cdot)$
are two functions implementing the detailed context-modulation,
where the input global context plays an essential role
as reference information.


\subsection{Global Context as Reference Information}

\begin{figure}[t]
\begin{center}
\includegraphics[scale=0.34]{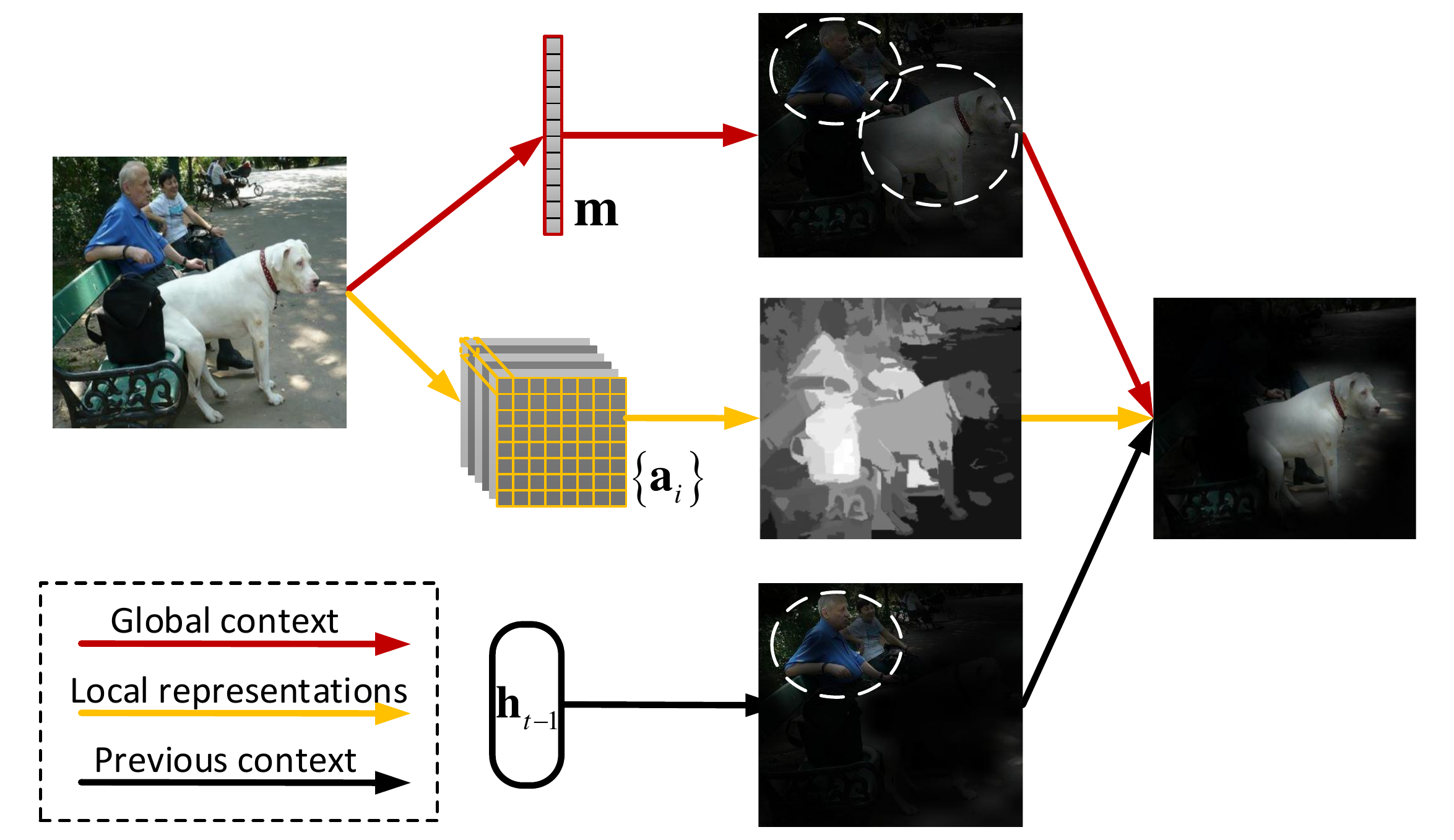}
\end{center}
\caption{Illustration of context-modulated attention (the lighter areas indicate the attended instances, best
viewed in colors).}
\label{fig:context}
\end{figure}

To illustrate the details of the context-modulated attention,
we take an image for example in Figure \ref{fig:context},
the case for sentence is similar.
The global feature \textbf{m} provides a statistical summary of the image scene,
including semantic instances and their relationship with each other.
Such a summary can not only provide reference information about
expected instances, e.g., man and dog,
but also cause the perception of one instance to generate
strong expectations about other instances \cite{chun1999top}.
The local representations $\left\{ {{{\textbf{a}}_i}}| {{\textbf{a}}_i} \in {\mathbb{R}^{F}}  \right\}_{i=1,\cdots,I}$
describe all the divided regions independently
and are used to compute the initial saliency map.
The hidden state at the previous timestep ${\textbf{h}_{t - 1}}$
indicates the already attended instances in the image, e.g., man.

To select which instance to attend to next,
the attention scheme should first refer to the global context to find an instance,
and then compare it with previous context to see if this instance has already been attended to.
If yes (e.g., selecting the man), the scheme will refer to the global context again to find another instance.
Otherwise (e.g., selecting the dog), regions in the initial saliency map corresponding to the instance will be highlighted.
For efficient implementation, we simulate such a context-modulated attentional procedure using
a simple three-way multilayer perceptrons (MLP)
as follows:
\begin{equation} \label{eqn:e2}
\begin{aligned}
{f_{{p}}}(\textbf{m}, {\textbf{a}_i}, {{\textbf{h}}_{t - 1}}) & =  {\kern 0pt} {\textbf{w}_p}(\sigma (\textbf{m}{W_\textbf{m}} + {\textbf{b}_\textbf{m}}) + \sigma ({\textbf{a}_i}{W_\textbf{a}} + {\textbf{b}_\textbf{a}}) \\
&+ \sigma ({\textbf{h}_{t - 1}}{W_\textbf{h}} + {\textbf{b}_\textbf{h}})) + {{b}_p}
\end{aligned}
\end{equation}
where $\sigma$ denotes the sigmoid activation function. ${\textbf{w}_p}$ and ${{b}_p}$ are a weight vector and a scalar bias.
Note that in this equation,
the information in initial saliency map is additively modulated by
global context and subtractively modulated by previous context,
to finally produce the instance-aware saliency map.

The attention schemes in \cite{xu2015show,bahdanau2014neural,ba2014multiple}
consider only previous context without global context at each timestep,
they have to alternatively use step-wise labels serving as expected instance
information to guide the attentional procedure.
But such strong supervision can only be available for
limited tasks, e.g., the sequential words of sentence for image caption \cite{xu2015show},
and multiple class labels for multi-object recognition \cite{ba2014multiple}.
For image and sentence matching, the words of sentence cannot be used as
supervision information since we also have to select salient instances from
the sentence to match image instances.
In fact, we perform experiments without using global context in Section \ref{sect:gc},
but find that some instances like man and dog cannot be well attended to.
It mainly results from the reason that without global context, the attention scheme
can only refer to the initial saliency map to select which instance to attend to next,
but the initial saliency map is computed from local representations
that contain little instance information as well as relationship among instances.

\subsection{Similarity Measurement and Aggregation}


According to the predicted pairwise instance-aware saliency maps,
we compute the weighted sum representations ${ \textbf{a}'_{t}}$ and ${ \textbf{w}'_{t}}$
to adaptively describe the attended image and sentence instances, respectively.
We sum all the products of element-wise multiplication
between each local representation (e.g., ${{\textbf{a}}_i}$)
and its corresponding saliency value (e.g., ${p_{t,i}}$):
\begin{equation} \label{eqn:e3}
{\textbf{a}'_t} = \sum\nolimits_{i = 1}^I {{p_{t,i}}{\textbf{a}_i}},  {\kern 7pt}
{\textbf{w}'_t} = \sum\nolimits_{j = 1}^J {{q_{t,j}}{\textbf{w}_j}}
\end{equation}
where instance candidates with higher saliency values
contribute more to the instance representations.
Then, to measure the local similarity of the attended pairwise instances at the $t$-th timestep,
we jointly feed their obtained representations
${ \textbf{a}'_{t}}$ and ${ \textbf{w}'_{t}}$ into a two-way MLP, and
regard the output $\textbf{s}_t$ as the representation of the local similarity,
as shown in Figure \ref{fig:details} (c).

From the 1-st to $T$-th timestep, we obtain a sequence of representations
of local similarities $\left\{ \textbf{s}_t \right\}_{t=1,\cdots,T}$,
where $T$ is the total number of timesteps.
To aggregate these local similarities for the global similarity,
we use a LSTM network to sequentially take them as inputs,
where the hidden states
$\left\{ \textbf{h}_t \in {\mathbb{R}^{H}} \right\}_{t=1,\cdots,T}$
dynamically propagate the captured local similarities
until the end.
The LSTM includes various gate mechanisms including
memory state ${\textbf{c}_t}$, hidden state ${\textbf{h}_t}$,
input gate $\textbf{i}_{t}$, forget gate $\textbf{f}_{t}$
and output gate $\textbf{o}_{t}$,
which can well suit the complex nature of similarity aggregation:
\begin{equation} \label{eqn:e5}
\begin{aligned}
&{\textbf{i}_t} = \sigma ({W_{\textbf{s}\textbf{i}}}{\textbf{s}_t} + {W_{\textbf{h}\textbf{i}}}{\textbf{h}_{t - 1}} + {\textbf{b}_\textbf{i}}), \\
&{\textbf{f}_t} = \sigma ({{W}_{\textbf{s}\textbf{f}}}{\textbf{s}_t} + {W_{\textbf{h}\textbf{f}}}{\textbf{h}_{t - 1}} + {\textbf{b}_\textbf{f}}),\\
&{\textbf{c}_t} = {\textbf{f}_t}\odot{\textbf{c}_{t - 1}} + {\textbf{i}_t}\odot\tanh ({W_{\textbf{s}\textbf{c}}}{\textbf{s}_t} + {W_{\textbf{h}\textbf{c}}}{\textbf{h}_{t - 1}} + {\textbf{b}_\textbf{c}}),\\
&{\textbf{o}_t} = \sigma ({W_{\textbf{s}\textbf{o}}}{\textbf{s}_t} + {W_{\textbf{h}\textbf{o}}}{\textbf{h}_{t - 1}} + {\textbf{b}_\textbf{o}}),
{\textbf{h}_t} = {\textbf{o}_t}\odot\tanh ({\textbf{c}_t})
\end{aligned}
\end{equation}
where $\odot$ denotes element-wise multiplication.

The hidden state at the last timestep ${\textbf{h}_{T}}$
can be regarded as the aggregated representation of all the local similarities.
We use a MLP that takes ${\textbf{h}_{T}}$ as the input
and produces the final matching score $s$ as global similarity:
\begin{equation} \label{eqn:e6}
\begin{aligned}
s = {\textbf{w}_{\textbf{h}s}}\left( {\sigma \left( {{{W}_{\textbf{hh}}}{\textbf{h}_t} + {\textbf{b}_\textbf{h}}} \right)} \right) + {b_s}.
\end{aligned}
\end{equation}


\subsection{Model Learning}

The proposed sm-LSTM can be trained with a structured objective function
that encourages the matching scores of matched images and sentences to be larger
than those of mismatched ones:
\begin{equation} \label{eqn:e5}
\begin{aligned}
\sum\nolimits_{ik}{{\max \left\{ {0,m - s_{ii}+ s_{ik}} \right\}}}  + {{\max \left\{ {0,m - s_{ii} + s_{ki}} \right\}} }
\end{aligned}
\end{equation}
where $m$ is a tuning parameter,
and $s_{ii}$ is the score of matched $i$-th image and $i$-th sentence.
$s_{ik}$ is the score of mismatched $i$-th image and $k$-th sentence,
and vice-versa with $s_{ki}$.
We empirically set the total number of mismatched pairs for each matched pair
as 100 in our experiments.
Since all modules of our model including the extraction of local representations
and global contexts can constitute a whole deep network,
our model can be trained in an end-to-end manner from raw image and sentence to matching score,
without pre-/post-processing.
For efficient optimization, we fix the weights of CNN and use pretrained weights
as stated in Section \ref{sent:details}.

In addition, we add a pairwise doubly stochastic regularization to the objective,
by constraining the sum of saliency values of any instance candidates at all timesteps to be 1:
\begin{equation} \label{eqn:e5}
\lambda \left( \sum\nolimits_{i} {( {1 - \sum\nolimits_{t} {{p_{t,i}}} } )}
+ \sum\nolimits_{j} {( {1 - \sum\nolimits_{t} {{q_{t,j}}} } )} \right)
\end{equation}
where $\lambda$ is a balancing parameter.
By adding this constraint, the loss will be large
when our model attends to the same instance for multiple times.
Therefore, it encourages the model to pay equal attention to
every instance rather than a certain one for information maximization.
In our experiments, we find that using this regularization can further improve the performance.

\begin{table*}[t] \small
\centering
\caption{Comparison results of image annotation and retrieval on the Flickr30K dataset. ($*$ indicates the ensemble or multi-model methods,
and $^{\dag}$ indicates using external text corpora or manual annotations.)}
\begin{tabular}{l|cccc|cccc|c}
\hline
\hline
\multirow{2}{0.7cm}{Method}     &  \multicolumn{4}{c|}{Image Annotation}  &  \multicolumn{4}{c|}{Image Retrieval} & \multirow{2}{0.7cm}{\textbf{Sum}}  \\
\cline{2-9}
     & R$@$1 & R$@$5  & R$@$10  & Med $r$  & R$@$1 & R$@$5  & R$@$10  & Med $r$ &    \\
\hline

RVP (T+I) \cite{chen2015mind}        &12.1 &27.8 &47.8 &11 &12.7 &33.1 &44.9 &12.5 & 178.4\\
Deep Fragment \cite{karpathy2014deep}  &14.2 &37.7 &51.3 &10 &10.2 &30.8 &44.2 &14 & 188.4\\
DCCA \cite{yan2015deep}           &16.7 &39.3 &52.9 &8 &12.6 &31.0 &43.0 &15 & 195.5\\
NIC \cite{vinyals2015show}           &17.0 &- &56.0 &7 &17.0 &- &57.0 &7 & -\\
DVSA (BRNN) \cite{karpathy2014vsa}    &22.2 &48.2 &61.4 &4.8 &15.2 &37.7 &50.5 &9.2 & 235.2\\
MNLM \cite{kiros2014unifying}        &23.0 &50.7 &62.9 &5 &16.8 &42.0 &56.5 &8 & 251.9\\
LRCN \cite{donahue2015long}            &- &- &- &- &17.5 &40.3 &50.8 &9 & -\\

m-RNN \cite{mao2014explain}       &35.4 &63.8 &73.7 &{3} &22.8 &50.7 &63.1 &{5} & 309.5\\
FV$^{\dag*}$ \cite{klein2015associating} &35.0 &62.0 &73.8 &{3} &25.0 &52.7 &66.0 &{5} & 314.5\\
m-CNN$^*$ \cite{ma2015multimodal}          &33.6 &{64.1} &{74.9} &{3} &{26.2} &{56.3} &{69.6} &{4} & 324.7\\
RTP+FV$^{\dag*}$  \cite{plummer2015flickr30k} &{37.4} &63.1 &74.3 &- &26.0 &{56.0} &{69.3} &- & 326.1\\
RNN+FV$^{\dag}$ \cite{lev2015rnn}          &34.7 &62.7 &72.6 &{3} &{26.2} &55.1 &69.2 &{4} & 320.5\\
DSPE+FV$^{\dag}$ \cite{wang2015learning}       &40.3 &68.9 &79.9 &- &29.7 &60.1 &72.1 &- & 351.0\\
\hline
\bf{Ours}:      & & & & & & & &\\
\hspace{0mm} sm-LSTM-mean                &25.9 &53.1 &65.4 &5 &18.1 &43.3 &55.7 &8 & 261.5\\
\hspace{0mm} sm-LSTM-att                &27.0 &53.6 &65.6 &5 &20.4 &46.4 &58.1 &7 & 271.1\\
\hspace{0mm} sm-LSTM-ctx                 &33.5 &60.6 &70.8 &3 &23.6 &50.4 &61.3 &5 &300.1\\
\hspace{0mm} sm-LSTM                  &42.4 &67.5 &79.9 &\bf{2} &28.2 &57.0 &68.4 &4 &343.4 \\
\hspace{0mm} sm-LSTM$^*$                 &\bf{42.5} &\bf{71.9} &\bf{81.5} &\bf{2} &\bf{30.2} &\bf{60.4} &\bf{72.3} &\bf{3} & \bf{358.7}\\


\hline
\hline
\end{tabular}

\label{table:f30k}
\end{table*}

\section{Experimental Results}
To demonstrate the effectiveness of the proposed sm-LSTM,
we perform experiments in terms of image annotation
and retrieval on two publicly available datasets.

\subsection{Datasets and Protocols}

The two evaluation datasets and their corresponding experimental protocols are described as follows.
1) \textbf{Flickr30k} \cite{young2014image}
consists of 31783 images collected from the Flickr website.
Each image is accompanied with 5 human annotated sentences.
We use the public training, validation and testing splits \cite{kiros2014unifying}, which
contain 28000, 1000 and 1000 images, respectively.
2) \textbf{Microsoft COCO} \cite{lin2014microsoft} consists of 82783
training and 40504 validation images, each of which is associated with 5 sentences.
We use the public training, validation and testing splits \cite{kiros2014unifying}, with
82783, 4000 and 1000 images, respectively.

\subsection{Implementation Details} \label{sent:details}
The commonly used evaluation criterions for image annotation and retrieval
are ``$\rm R@1$'', ``$\rm R@5$'' and ``$\rm R@10$'',
i.e., recall rates at the top 1, 5 and 10 results.
Another one is ``Med r'' which is the median rank of the first ground truth result.
We compute an additional criterion ``Sum'' to evaluate the overall performance for
both image annotation and retrieval as follows:
\begin{equation*}
\rm Sum = \underbrace {\rm R@1 + \rm R@5 + \rm R@10}_{ Image{\kern 3pt}  annotation} + \underbrace {\rm R@1 + \rm R@5 + \rm R@10}_{Image {\kern 3pt} retrieval}
\end{equation*}

To systematically validate the effectiveness,
we experiment with five variants of sm-LSTMs:
(1) sm-LSTM-mean does not use the attention scheme to obtain weighted sum
representations for selected instances but instead directly uses mean vectors,
(2) sm-LSTM-att only uses the attention scheme but does not exploit global context,
(3) sm-LSTM-ctx does not use the attention scheme but only exploits global context,
(4) sm-LSTM is our full model that uses both the attention scheme and global context,
and (5) sm-LSTM$^*$ is an ensemble of the described four models above,
by summing their cross-modal similarity matrices together in a similar way as \cite{ma2015multimodal}.

We use the 19-layer VGG network \cite{simonyan2014very} to initialize
our CNN to extract 512 feature maps
(with a size of 14$\times$14) in ``conv5-4'' layer
as representations for image instance candidates,
and a feature vector in ``fc7'' layer as the image global context.
We use MNLP \cite{kiros2014unifying} to initialize our sentence-based LSTM
and regard the hidden state at the last timestep as the sentence global context,
while our BLSTM for representing sentence candidates are directly learned from raw sentences with a dimension of hidden state as 512.
For image, the dimensions of local and global context features are $F$=$512$ and $D$=$4096$, respectively,
and the total number of local regions is $I$=$196$ (14$\times$14).
For sentence, the dimensions of local and global context features are $G$=$1024$ and $E$=$1024$, respectively.
We set the max length for all the sentences as 50, i.e., the number of split words $J$=$50$,
and use zero-padding when a sentence is not long enough.
Other parameters are empirically set as follows:
$H$=$1024$, $\lambda$=$100$, $T$=$3$ and $m$=$0.2$.

\begin{table*}[t] \small
\centering
\caption{Comparison results of image annotation and retrieval on the Microsoft COCO dataset. ($*$ indicates the ensemble or multi-model methods,
and $^{\dag}$ indicates using external text corpora or manual annotations.)}
\begin{tabular}{l|cccc|cccc|c}
\hline
\hline
\multirow{2}{0.7cm}{Method}     &  \multicolumn{4}{c|}{Image Annotation}  &  \multicolumn{4}{c|}{Image Retrieval} & \multirow{2}{0.7cm}{\textbf{Sum}}  \\
\cline{2-9}
     & R$@$1 & R$@$5  & R$@$10  & Med $r$  & R$@$1 & R$@$5  & R$@$10  & Med $r$ &    \\
\hline
STD$^{\dag*}$ \cite{kiros2015skip} &33.8 &67.7 &82.1 &3 &25.9 &60.0 &74.6 &4 & 344.1\\
m-RNN \cite{mao2014explain}           &41.0 &73.0 &83.5 &{2} &29.0 &42.2 &77.0 &{3} & 345.7\\
FV$^{\dag*}$ \cite{klein2015associating}     &39.4 &67.9 &80.9 &{2} &25.1 &59.8 &76.6 &4 & 349.7\\
DVSA \cite{karpathy2014vsa}               &38.4 &69.9 &80.5 &\bf{1} &27.4 &60.2 &74.8 &{3} & 351.2\\
MNLM \cite{kiros2014unifying}        &43.4 &75.7 &85.8 &{2} &31.0 &66.7 &79.9 &{3} & 382.5\\
m-CNN$^*$ \cite{ma2015multimodal}              &42.8 &73.1 &84.1 &{2} &32.6 &{68.6} &{82.8} &{3} & 384.0\\
RNN+FV$^{\dag}$ \cite{lev2015rnn}          &40.8 &71.9 &83.2 &2 &29.6 &64.8 &80.5 &3 & 370.8\\
OEM \cite{vendrov2015order}       &46.7 &- &88.9 &2 &37.9 &- &85.9 &\bf{2} &-\\
DSPE+FV$^{\dag}$ \cite{wang2015learning}       &50.1 &79.7 &89.2 &- &39.6 &75.2 &86.9 &- & 420.7\\

\hline
\bf{Ours}:      & & & & & & & &\\
\hspace{0mm} sm-LSTM-mean      &33.1 &65.3 &78.3 &3 &25.1 &57.9 &72.2 &4 &331.9 \\
\hspace{0mm} sm-LSTM-att      &36.7 &69.7 &80.8 &2 &{29.1} &{64.8} &{78.4} &3 &{359.5} \\
\hspace{0mm} sm-LSTM-ctx            &39.7 &70.2 &84.0 &2 &32.7 &68.1 &81.3 &3 &376.0\\
\hspace{0mm} sm-LSTM            &52.4 &81.7 &90.8 &\bf{1} &38.6 &73.4 &84.6 &\bf{2} &421.5\\
\hspace{0mm} sm-LSTM$^*$      &\bf{53.2} &\bf{83.1} &\bf{91.5} &\bf{1} &\bf{40.7} &\bf{75.8} &\bf{87.4} &\bf{2} & \bf{431.8}\\


\hline
\hline
\end{tabular}
\label{table:coco}
\end{table*}

\begin{figure*}[t]
\begin{center}
\includegraphics[scale=0.67]{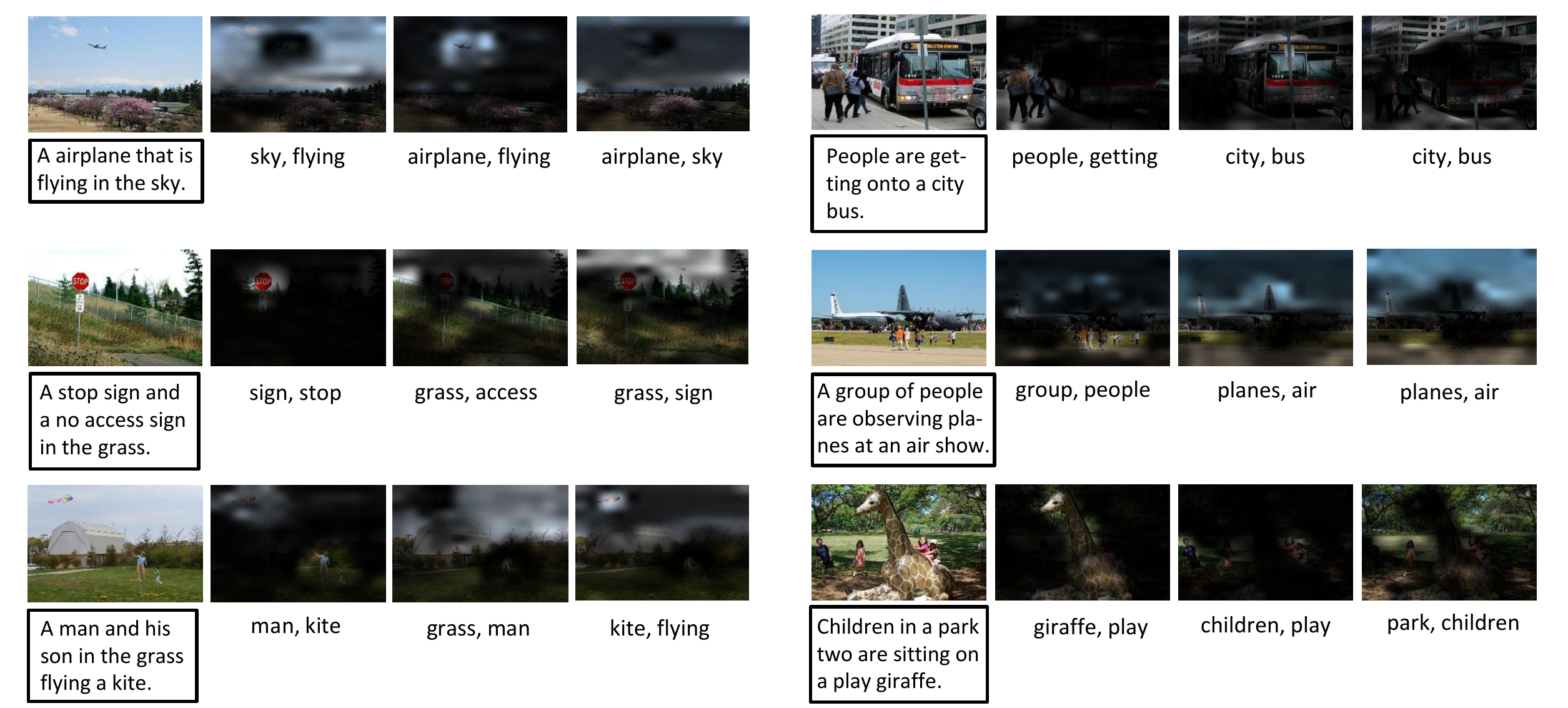}
\end{center}
\caption{Visualization of attended image and sentence instances at three different timesteps (best
viewed in colors).}
\label{fig:pairwise}
\end{figure*}

\subsection{Comparison with State-of-the-art Methods}

We compare sm-LSTMs with several recent state-of-the-art methods
on the Flickr30k and Microsoft COCO datasets
in Tables \ref{table:f30k} and \ref{table:coco}, respectively.
We can find that sm-LSTM$^*$ can achieve much better performance
than all the compared methods on both datasets.
Our best single model sm-LSTM outperforms the
state-of-the-art DSPE+FV$^{\dag}$ in image annotation, but performs slightly worse
than it in image retrieval.
Different from DSPE+FV$^{\dag}$ that uses external text corpora to learn discriminative sentence features, our model
learns them directly from scratch in an end-to-end manner.
Beside DSPE+FV$^{\dag}$, the sm-LSTM performs better than other compared methods by a large margin.
These observations demonstrate that dynamically selecting image-sentence instances and
aggregating their similarities is very effective for cross-modal retrieval.

When comparing among all the sm-LSTMs, we can conclude as follows.
1) Our attention scheme is effective,
since sm-LSTM-att consistently outperforms sm-LSTM-mean on both datasets.
When exploiting only context information without the attention scheme,
sm-LSTM-ctx achieves much worse results than sm-LSTM.
2) Using global context to modulate the attentional procedure
is very useful, since sm-LSTM greatly outperforms sm-LSTM-att
with respect to all evaluation criterions.
3) The ensemble of four sm-LSTM variants as sm-LSTM$^*$ can further improve the performance.

\begin{table}[t] \small
\centering
\caption{The impact of different numbers of timesteps on the Flick30k dataset. $T$: the number of timesteps in the sm-LSTM.}
\begin{tabular}{l|ccc|ccc}
\hline
\hline
\multirow{2}{0.7cm}{}     &  \multicolumn{3}{c|}{Image Annotation}  &  \multicolumn{3}{c}{Image Retrieval}   \\
\cline{2-7}
     & R$@$1 & R$@$5  & R$@$10   & R$@$1 & R$@$5  & R$@$10     \\
\hline
\hspace{0mm} $T=1$      &38.8 &65.7 &76.8 &28.0 &56.6 &68.2 \\
\hspace{0mm} $T=2$      &38.0 &\bf{68.9} &77.9 &28.1 &56.5 &68.1 \\
\hspace{0mm} $T=3$      &\bf{42.4} &67.5 &\bf{79.9} &\bf{28.2} &\bf{57.0} &\bf{68.4} \\
\hspace{0mm} $T=4$      &38.2 &67.6 &78.5 &27.5 &56.6 &68.0  \\
\hspace{0mm} $T=5$      &38.1 &68.2 &78.4 &28.1 &56.0 &67.9  \\

\hline
\hline
\end{tabular}
\label{table:step}
\end{table}

\subsection{Analysis on Number of Timesteps} \label{sec:temporal-step}
For a pair of image and sentence, we need to manually set the number of timesteps $T$ in sm-LSTM.
Ideally, $T$ should be equal to the number of salient pairwise instances appearing in the image and sentence.
Therefore, the sm-LSTM can separately attend to these pairwise instances
within $T$ steps to measure all the local similarities.
To investigate what is the optimal number of timesteps,
in the following, we gradually increase $T$ from 1 to 5, and
analyze the impact of different numbers of timesteps on the performance of sm-LSTM
in Table \ref{table:step}.

From the table we can observe that sm-LSTM
can achieve its best performance when the
number of timesteps is 3. It indicates that it can
capture all the local similarity information by iteratively visiting both image and sentence for 3 times.
Intuitively, most pairs of images and sentences usually contain approximately 3 associated instances.
Note that when $T$ becomes larger than 3, the performance slightly drops.
It results from the fact that an overly complex network tends to overfit training data
by paying attention to redundant instances at extra timesteps.

\begin{figure*}[t]
\addtolength{\tabcolsep}{-2pt}
\centering
\begin{tabular}{ccccccc}


\includegraphics[width = 0.87in]{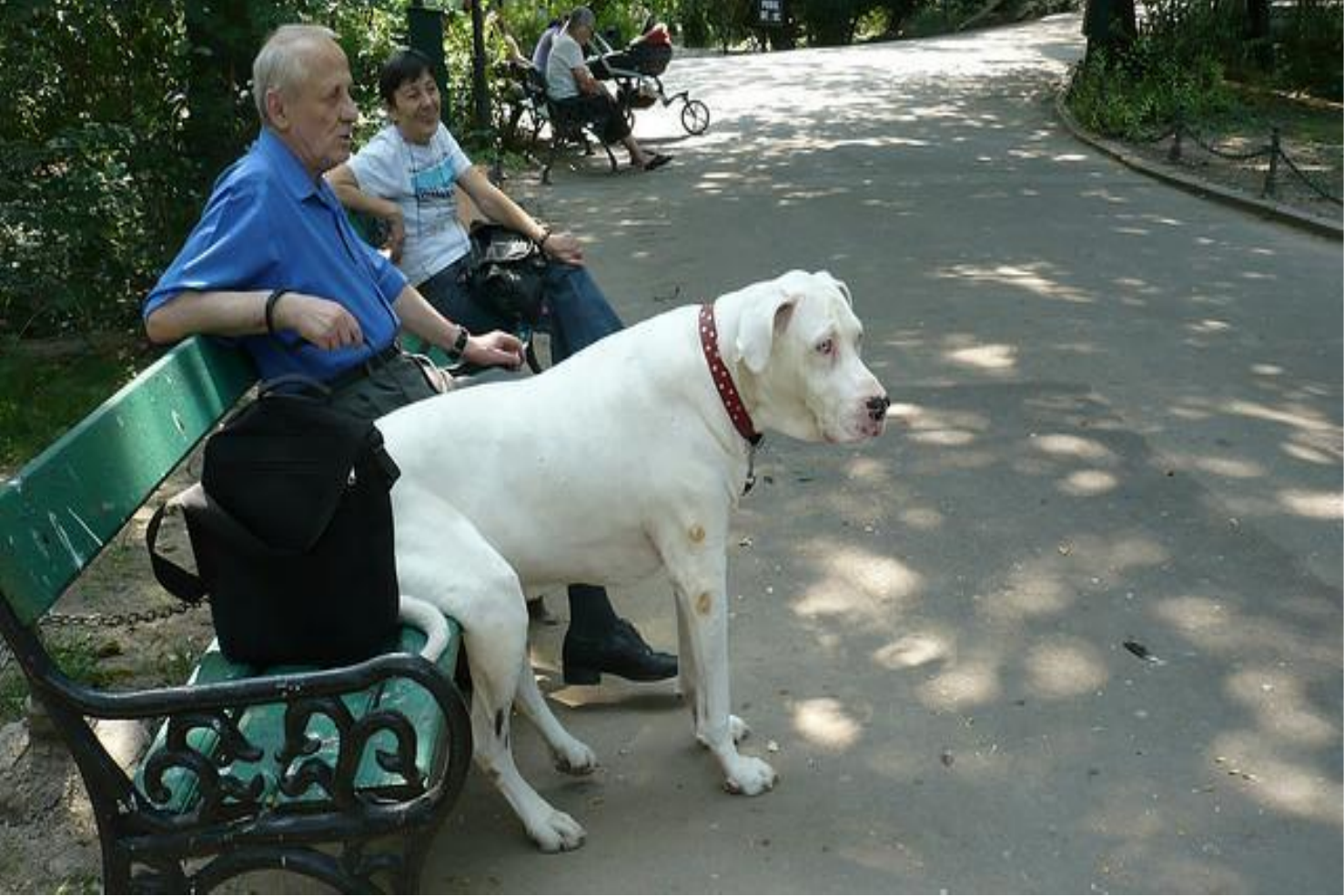}&
\includegraphics[width = 0.87in]{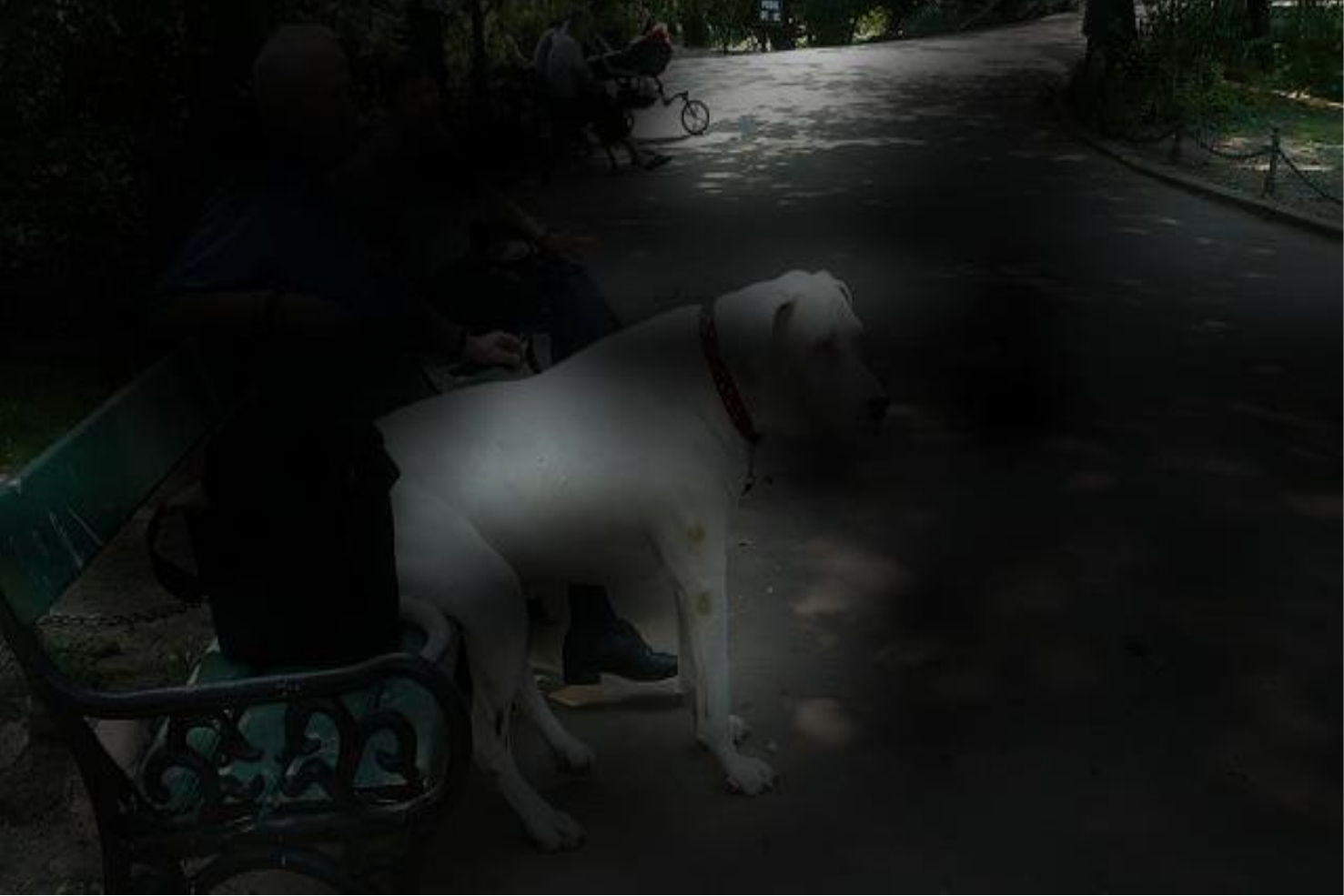}\hspace{-3mm}&\hspace{-3mm}
\includegraphics[width = 0.87in]{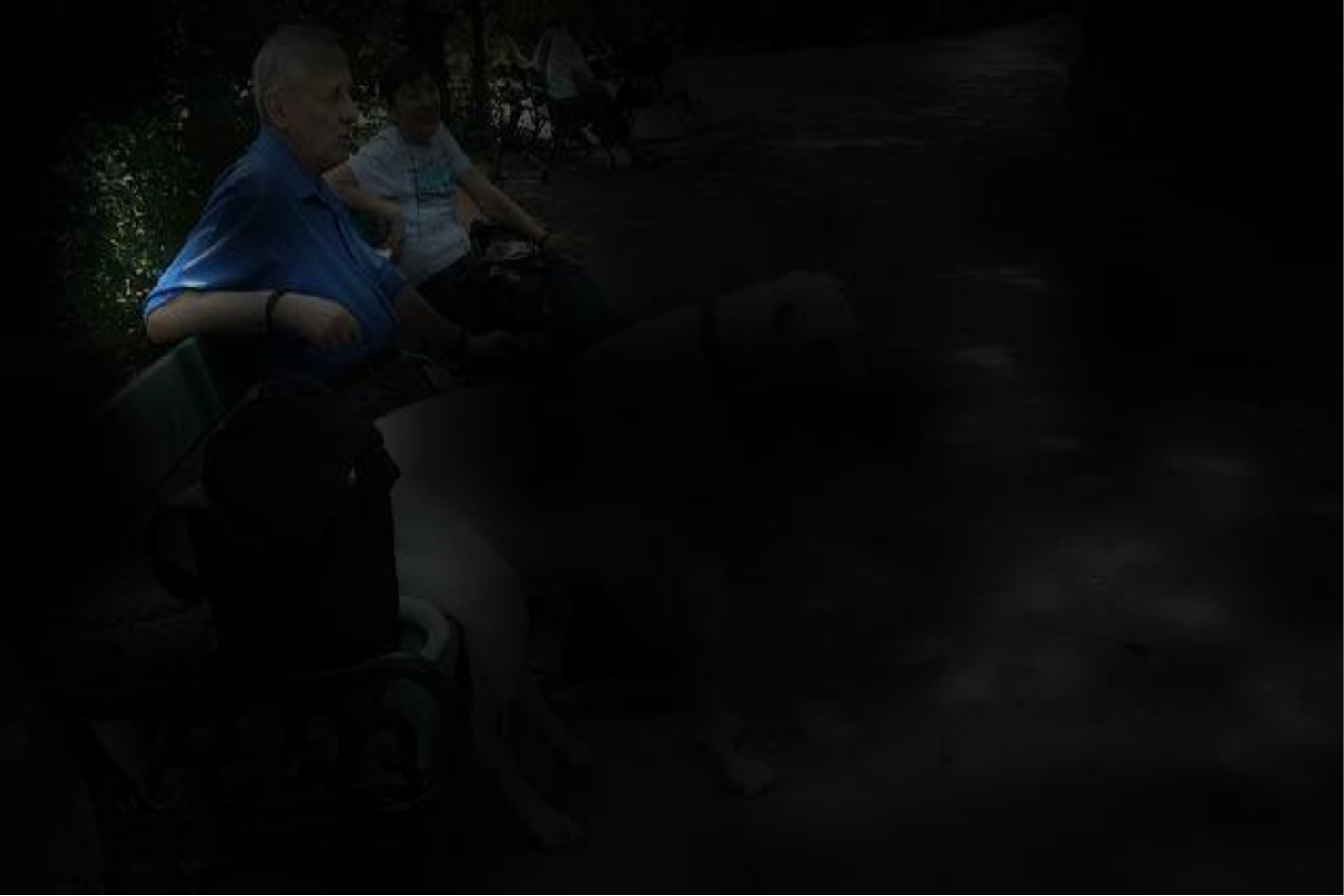}\hspace{-3mm}&\hspace{-3mm}
\includegraphics[width = 0.87in]{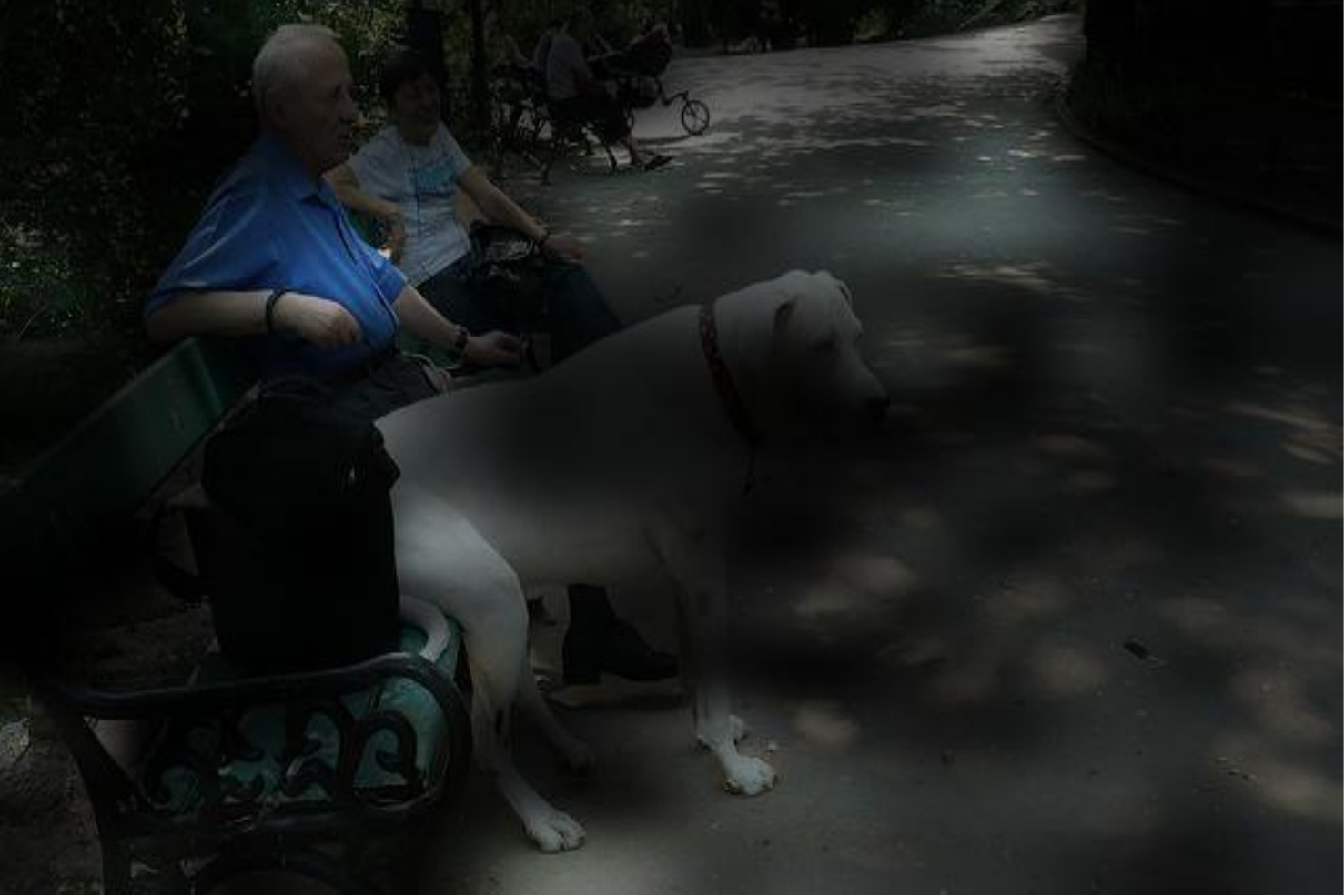}&
\includegraphics[width = 0.87in]{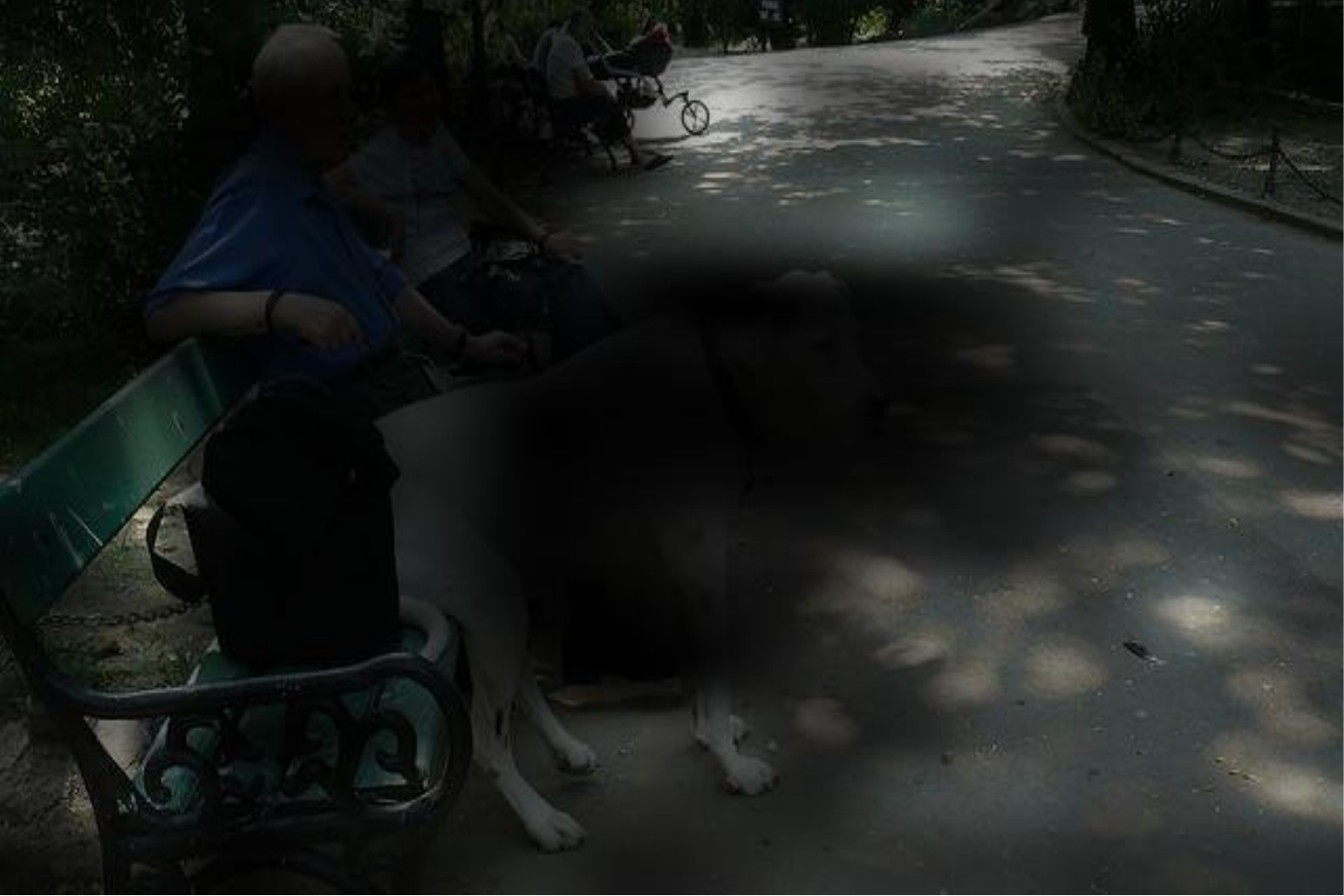}\hspace{-3mm}&\hspace{-3mm}
\includegraphics[width = 0.87in]{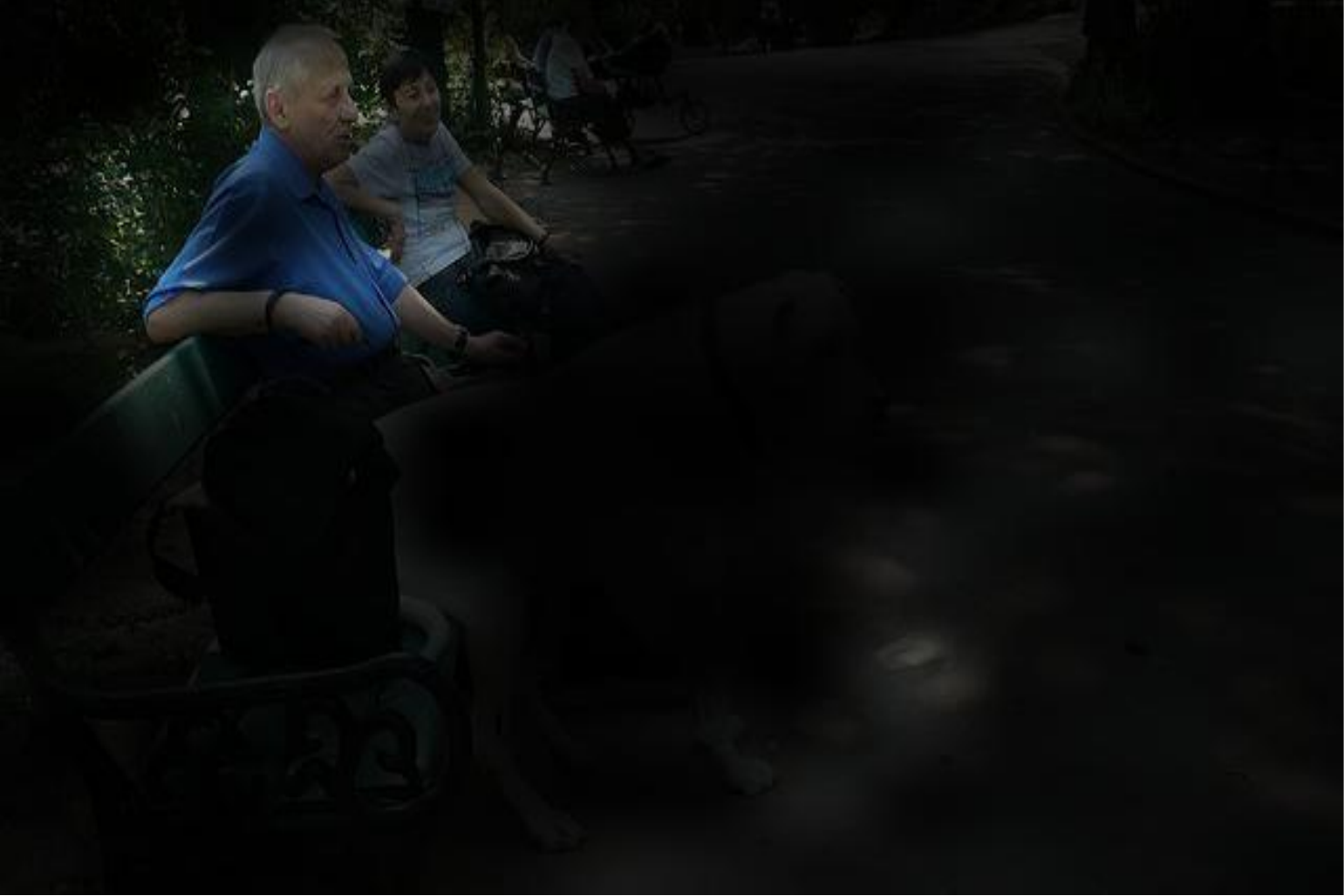}\hspace{-3mm}&\hspace{-3mm}
\includegraphics[width = 0.87in]{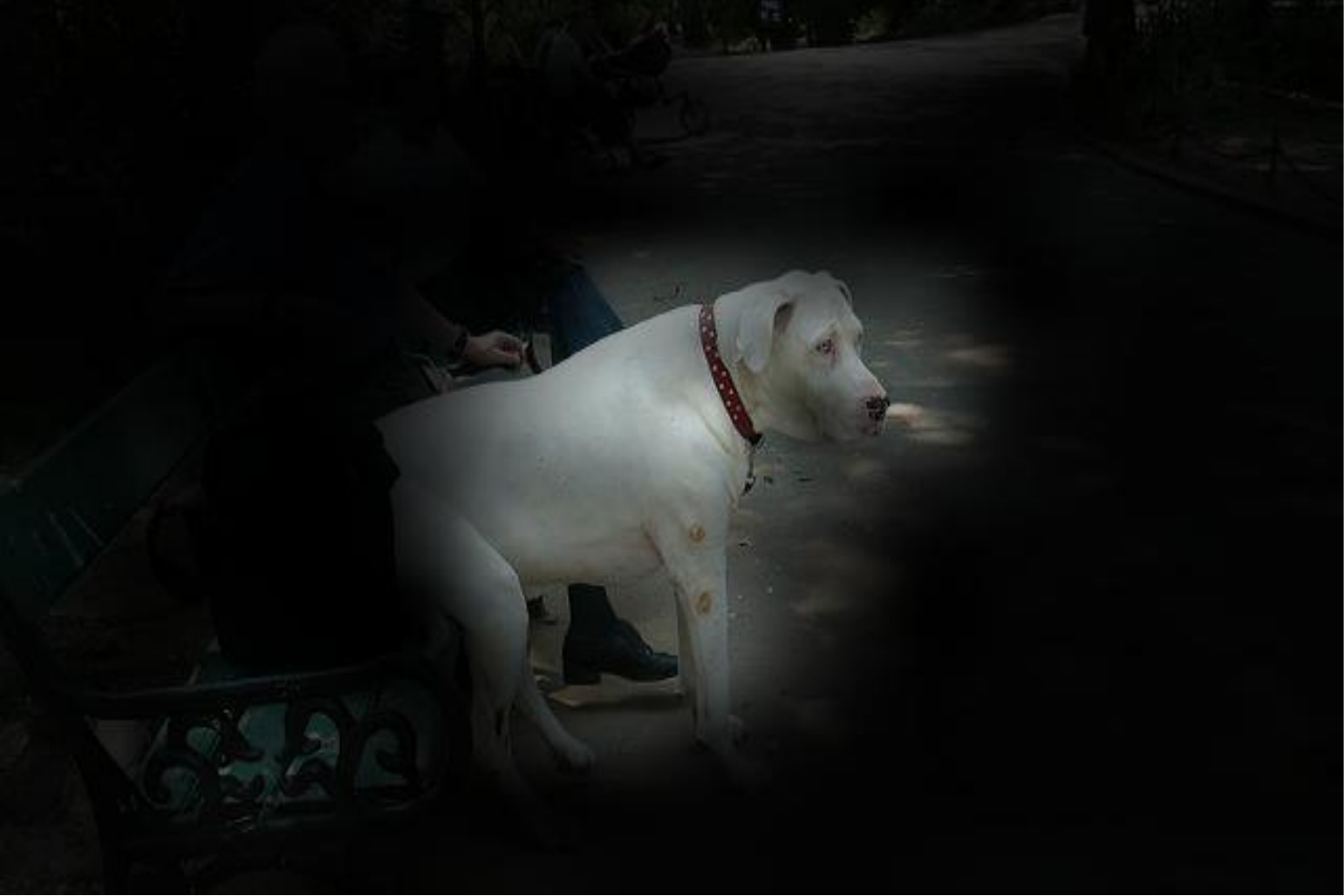}\\

\includegraphics[width = 0.87in]{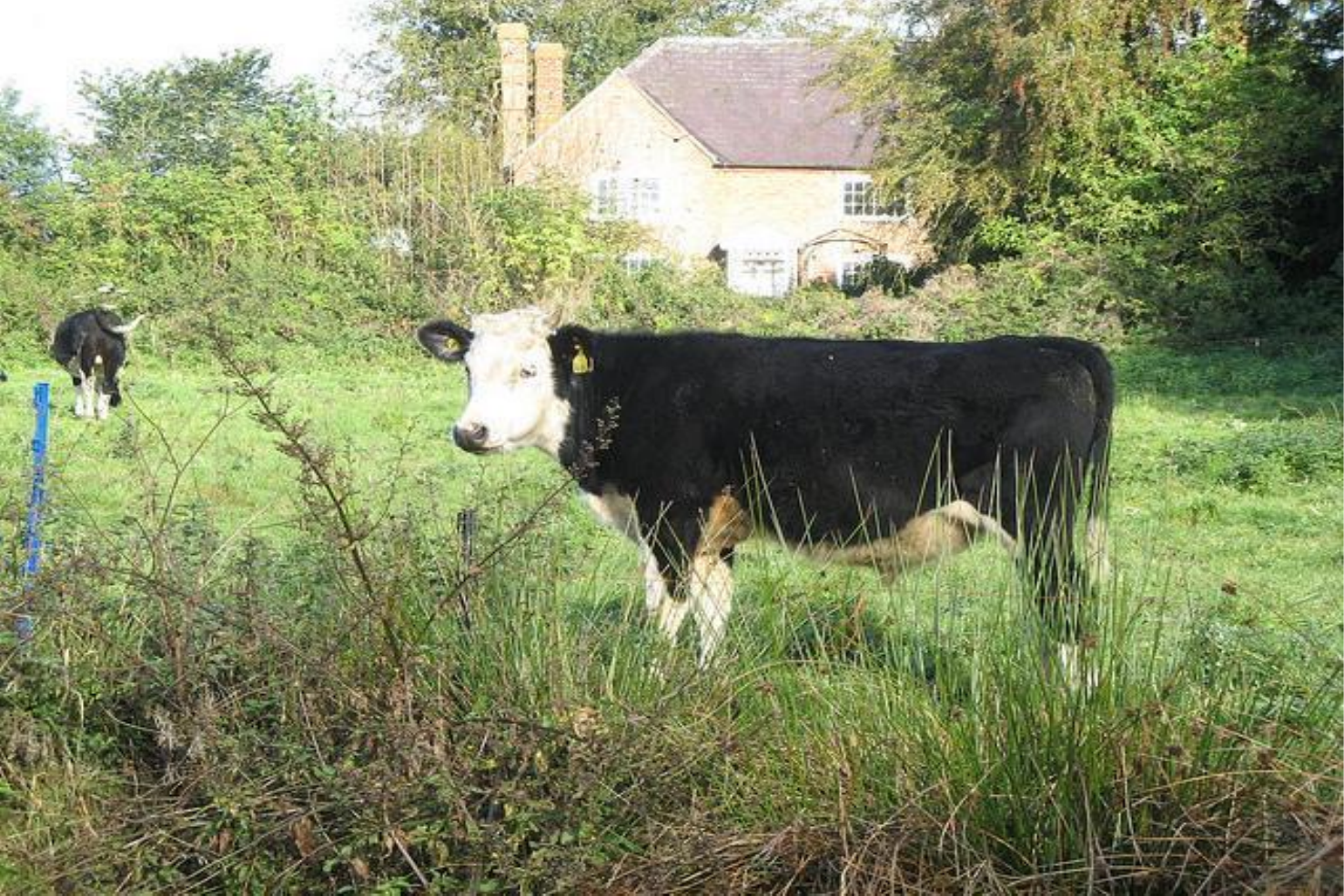}&
\includegraphics[width = 0.87in]{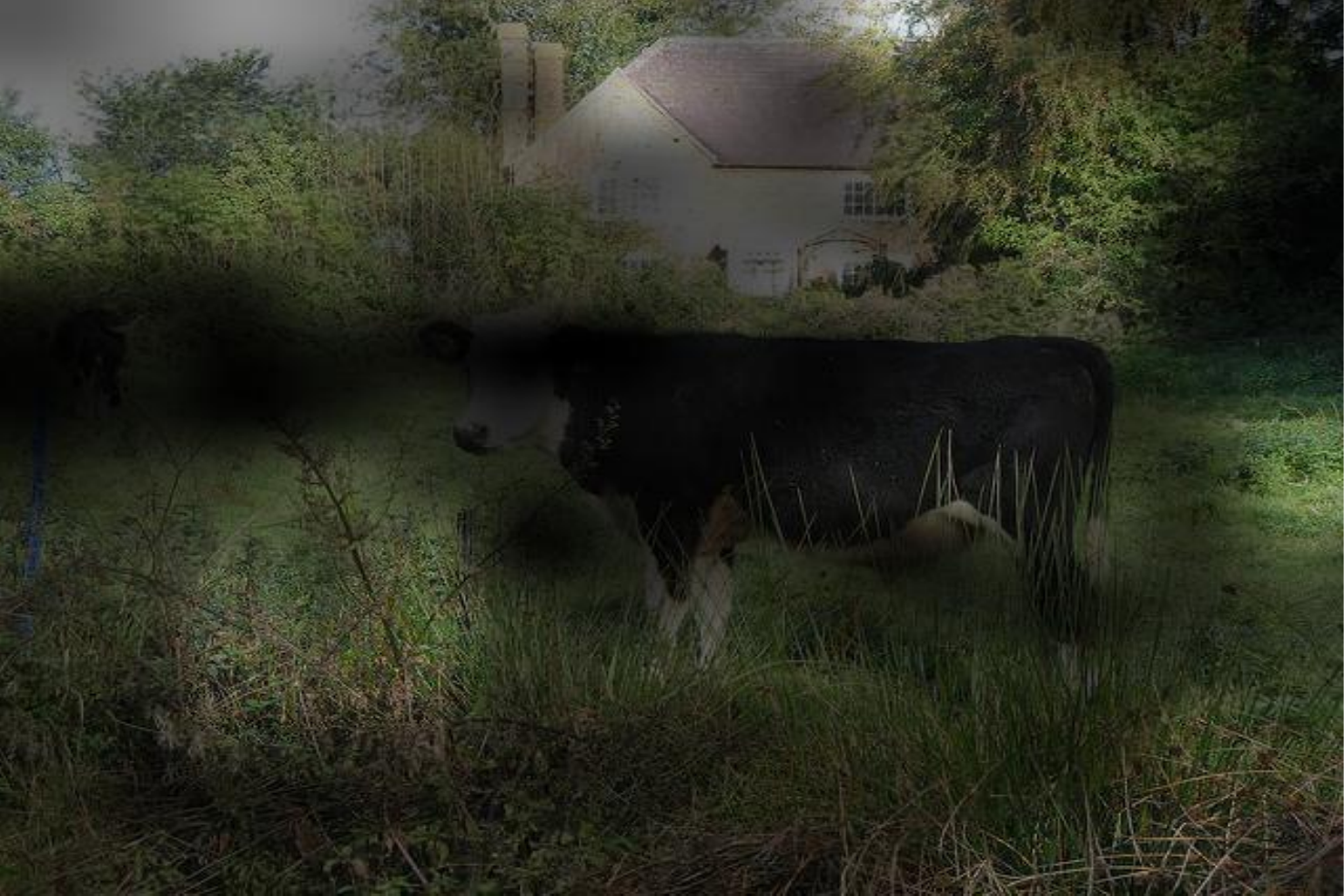}\hspace{-3mm}&\hspace{-3mm}
\includegraphics[width = 0.87in]{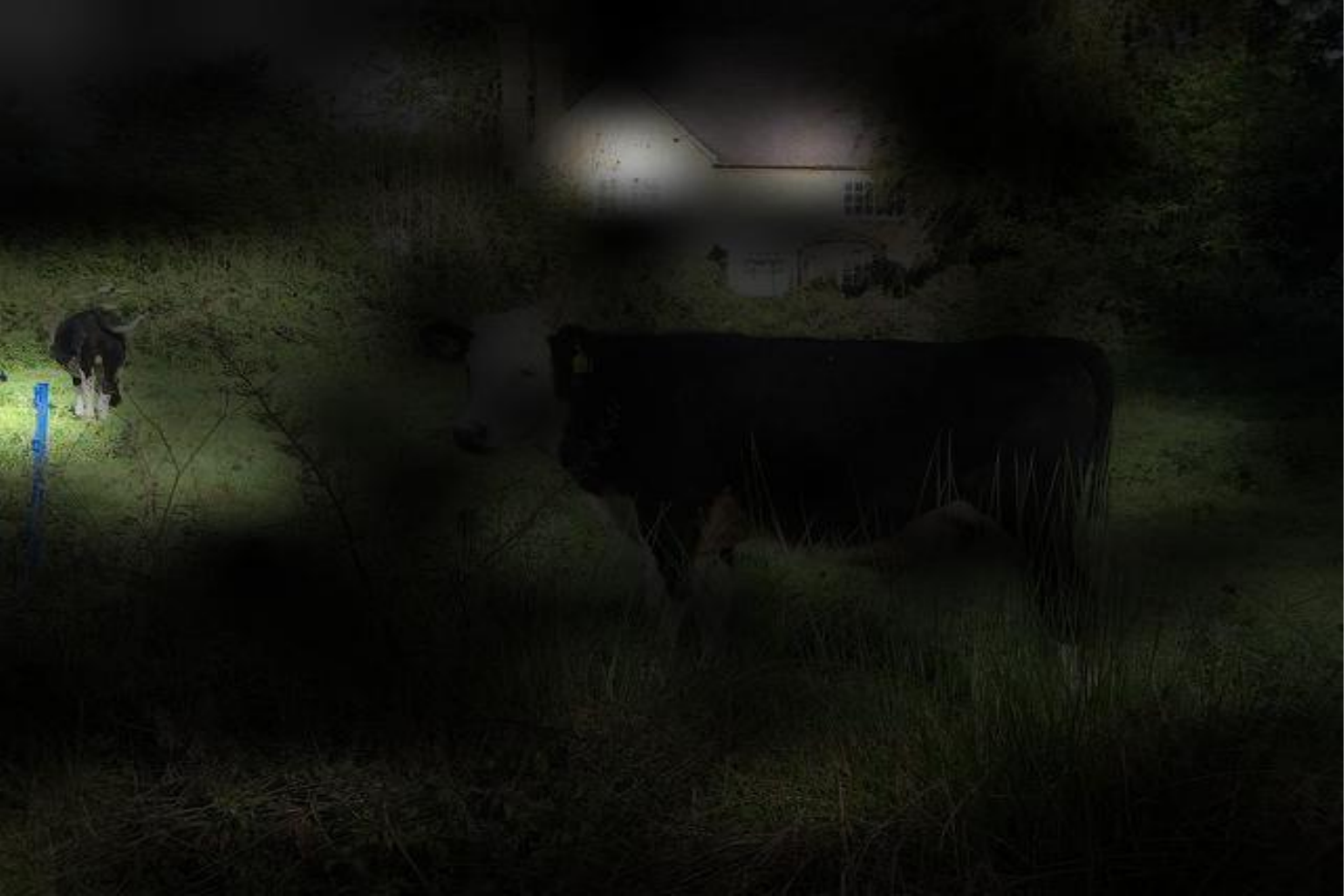}\hspace{-3mm}&\hspace{-3mm}
\includegraphics[width = 0.87in]{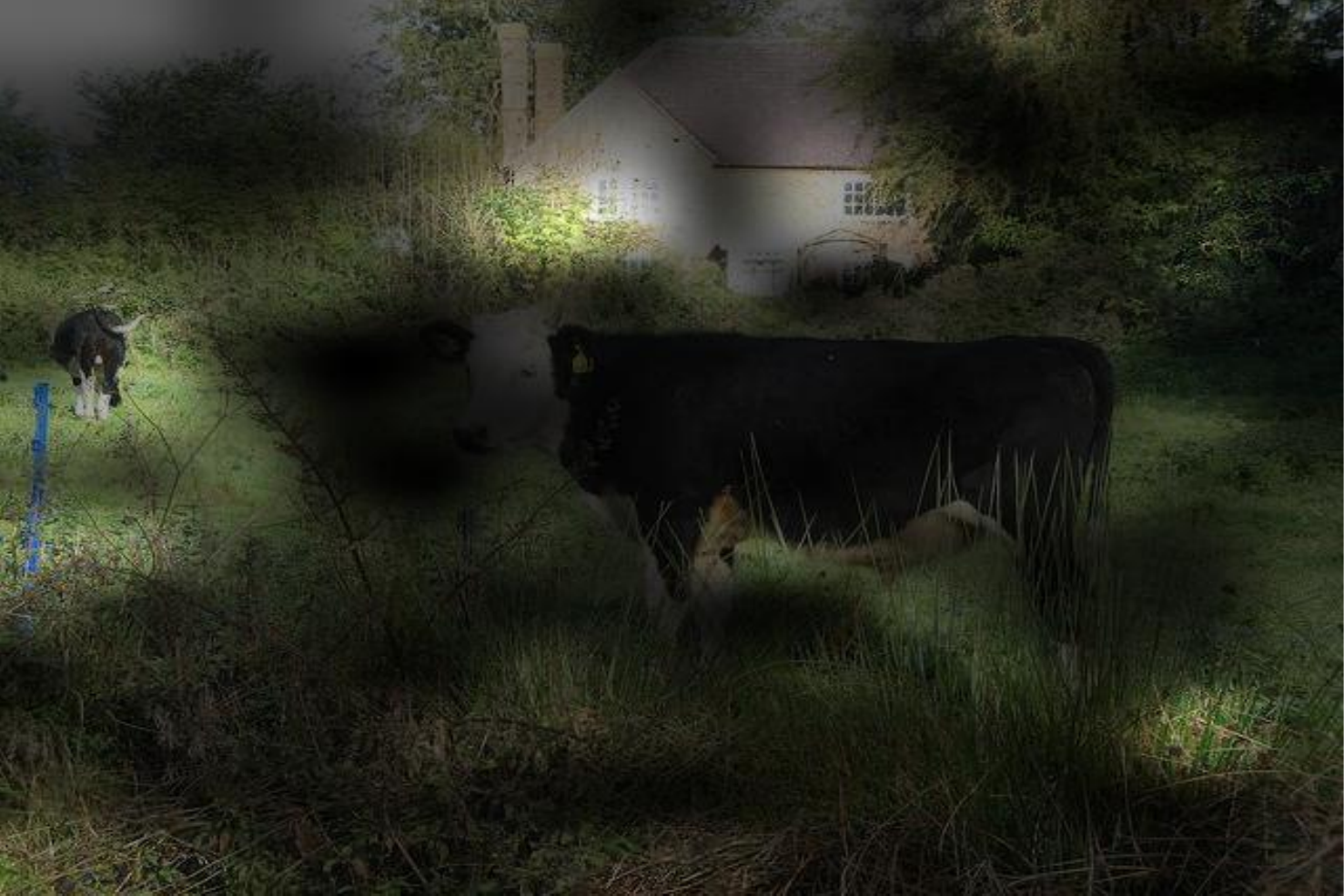}&
\includegraphics[width = 0.87in]{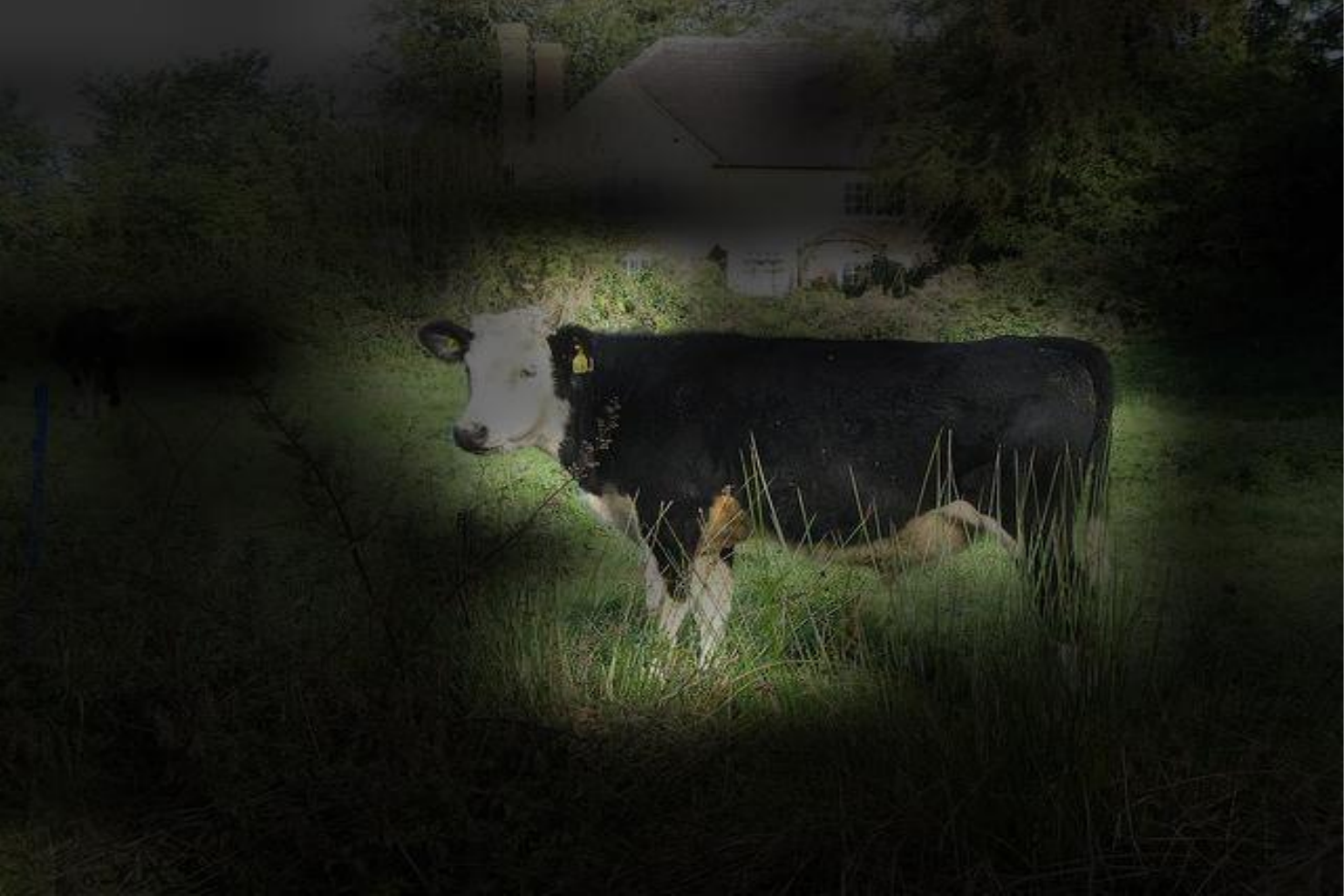}\hspace{-3mm}&\hspace{-3mm}
\includegraphics[width = 0.87in]{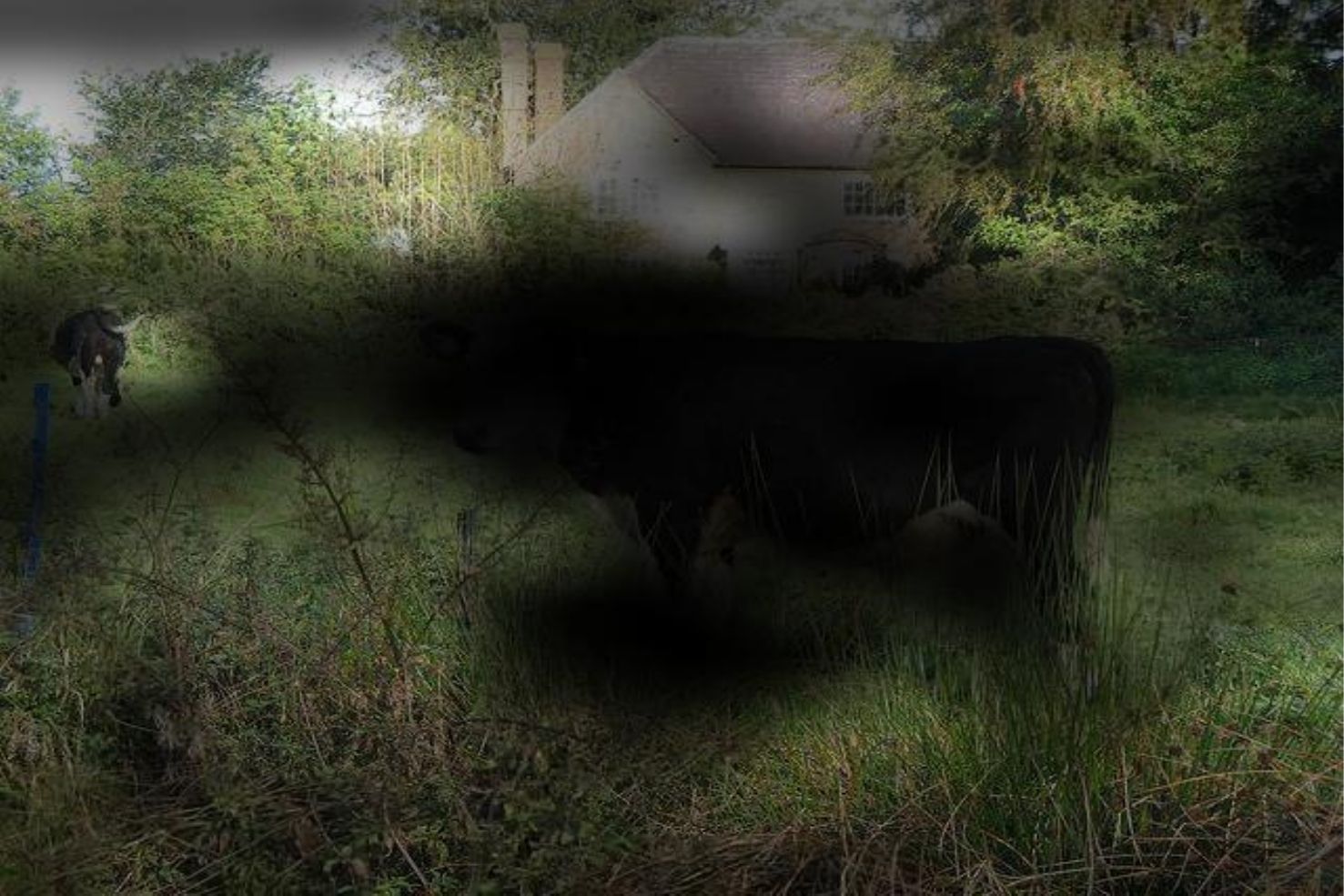}\hspace{-3mm}&\hspace{-3mm}
\includegraphics[width = 0.87in]{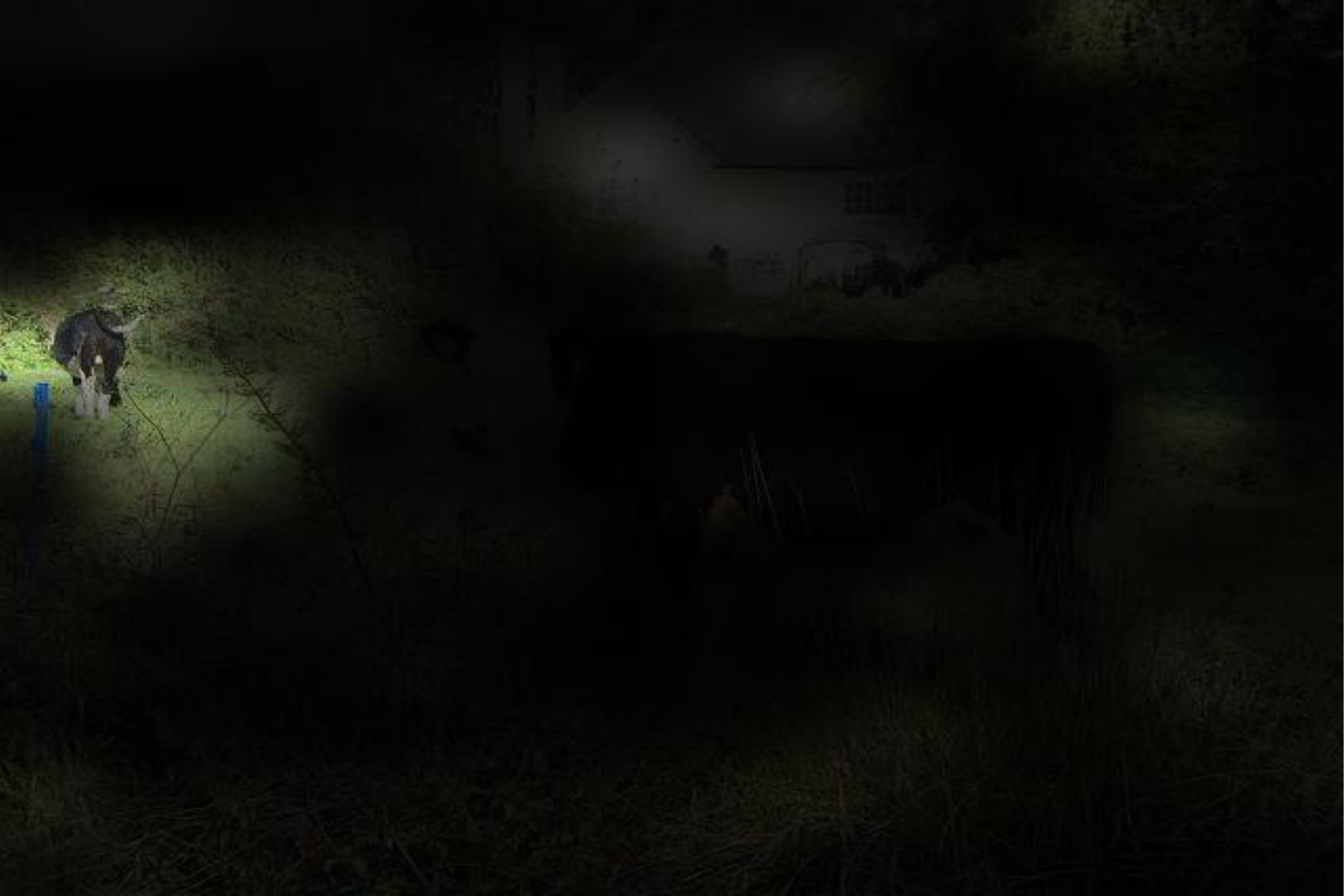}\\

(a) Input image&  \multicolumn{3}{c}{(b) Without global context (by sm-LSTM-att)} &  \multicolumn{3}{c}{(c) With global context (by sm-LSTM)} \\
\\
\end{tabular}
\caption{Attended image instances at three different timesteps, without or with global context, respectively (best
viewed in colors).}
\label{figure:map2}
\end{figure*}

\begin{table}[t] \small
\addtolength{\tabcolsep}{-0.5pt}
\centering
\caption{The impact of different values of the balancing parameter on the Flick30k dataset.
$\lambda$: the balancing parameter between structured objective and regularization term.}
\begin{tabular}{l|ccc|ccc}
\hline
\hline
\multirow{2}{0.7cm}{}     &  \multicolumn{3}{c|}{Image Annotation}  &  \multicolumn{3}{c}{Image Retrieval}   \\
\cline{2-7}
     & R$@$1 & R$@$5  & R$@$10   & R$@$1 & R$@$5  & R$@$10  \\
\hline
\hspace{0mm} $\lambda=0$      &37.9 &65.8 &77.7 &27.2 &55.4 &67.6 \\
\hspace{0mm} $\lambda=1$      &38.0 &66.2 &77.8 &27.4 &55.6 &67.7  \\
\hspace{0mm} $\lambda=10$      &38.4 &67.4 &77.7 &27.5 &56.1 &67.6 \\
\hspace{0mm} $\lambda=100$      &\bf{42.4} &\bf{67.5} &\bf{79.9} &\bf{28.2} &\bf{57.0} &\bf{68.4} \\
\hspace{0mm} $\lambda=1000$      &40.2 &67.1 &78.6 &27.8 &56.9 &67.9 \\

\hline
\hline
\end{tabular}
\label{table:R}
\end{table}

\subsection{Evaluation of Regularization Term}
In our experiments, we find that the proposed sm-LSTM
is inclined to focus on the same instance at all timesteps,
which might result from the fact that always selecting most informative
instances can largely avoid errors.
But it is not good for our model to comprehensively perceive the entire content
in the image and sentence.
So we add the pairwise doubly stochastic regularization term (in Equation \ref{eqn:e5})
to the structured objective,
with the aim to force the model to pay equal attention to all the potential instances at different locations.
We vary the values of balancing parameter $\lambda$ from 0 to 1000,
and compare the corresponding performance in Table \ref{table:R}.
From the table, we can find that the performance improves when $\lambda$$>$$0$,
which demonstrates the usefulness of paying attention to more instances.
In addition, when $\lambda$$=$$100$, the ms-LSTM can achieve the largest performance improvement,
especially for the task of image annotation.

\subsection{Visualization of Instance-aware Saliency Maps}

To verify whether the proposed model can selectively attend to salient
pairwise instances of image and sentence at different timesteps,
we visualize the predicted sequential instance-aware saliency maps by sm-LSTM,
as shown in Figure \ref{fig:pairwise}.
In particular for image, we resize the predicted saliency values at the $t$-th timestep
${\left\{ {p_{t,i}} \right\}}$ to
the same size as its corresponding original image, so that
each value in the resized map measures the importance of an image pixel at the same location.
We then perform element-wise multiplication between the resized saliency map and the original image
to obtain the final saliency map, where lighter areas indicate attended instances.
While for sentence, since different sentences have various lengths,
we simply present two selected words at each timestep
corresponding to the top-2 highest saliency values ${\left\{ {q_{t,j}} \right\}}$.

We can see that sm-LSTM can attend to different regions and words
at different timesteps in the images and sentences, respectively.
Most attended pairs of regions and words describe similar semantic concepts.
Taking the last pair of image and sentence for example, sm-LSTM sequentially focuses on
words: ``giraffe'', ``children'' and ``park'' at three different timesteps,
as well as the corresponding image regions referring to similar meanings.

\subsection{Usefulness of Global Context} \label{sect:gc}

To qualitatively validate the effectiveness of using global context,
we compare the resulting instance-aware saliency maps of images
generated by sm-LSTM-att and sm-LSTM in Figure \ref{figure:map2}.
Without the aid of global context, sm-LSTM-att cannot produce
accurate dynamical saliency maps as those of sm-LSTM.
In particular, it cannot well attend to semantically meaningful instances
such as ``dog'', ``cow'' and ``beach'' in the first,
second and third images, respectively.
In addition, sm-LSTM-att always finishes attending to salient instances within the first two steps,
and does not focus on meaningful instances at the third timestep any more.
Different from it, sm-LSTM focuses on more salient instances at all three timesteps.
These evidences show that global context modulation
can be helpful for more accurate instance selection.

\begin{figure}[t]
\addtolength{\tabcolsep}{-2pt}
\centering
\begin{tabular}{ccc}
\includegraphics[width = 0.8in]{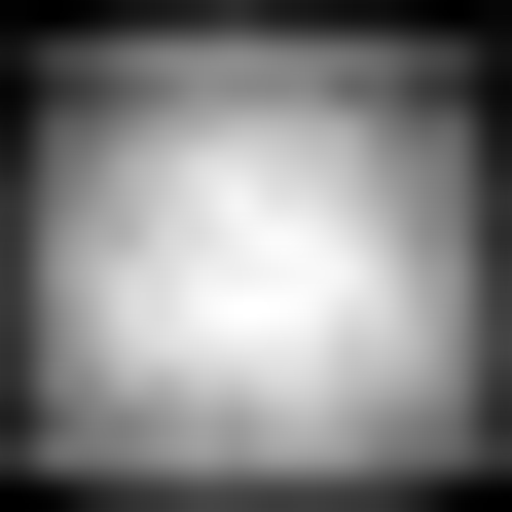} \hspace{2mm}&\hspace{2mm}
\includegraphics[width = 0.8in]{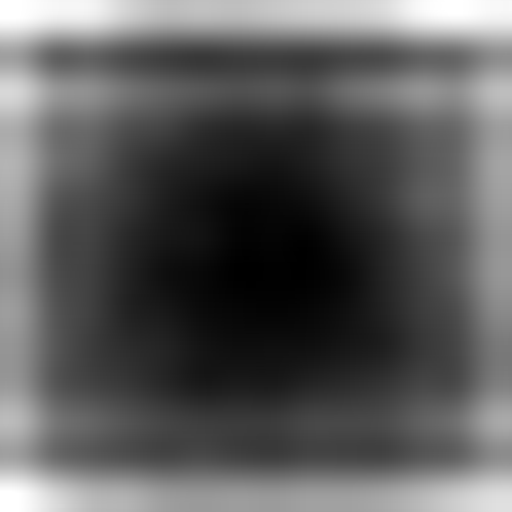} \hspace{2mm}&\hspace{2mm}
\includegraphics[width = 0.8in]{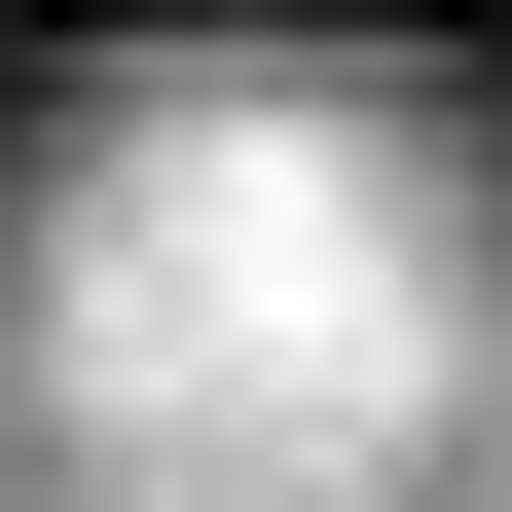}\\
(a) 1-st timestep  &  (b) 2-nd timestep &  (c) 3-rd timestep\\
\\
\end{tabular}
\caption{Averaged saliency maps at three different timesteps.}
\label{figure:mean}
\end{figure}

In Figure \ref{figure:mean}, we also compute the averaged saliency maps
(rescaled to the same size of 500$\times$500) for all the test images
at three different timesteps by sm-LSTM.
We can see that the proposed sm-LSTM statistically tends to focus on
the central regions at the first timestep,
which is in consistent with the observation of ``center-bias'' in
human visual attention studies \cite{tseng2009quantifying,bindemann2010scene}.
It is mainly attributed to the fact that salient instances mostly
appear in the cental regions of images.
Note that the model also attends to surrounding and lower regions
at the other two timesteps, with the goal to find various instances at different locations.


\section{Conclusions}
In this paper, we have proposed the selective multimodal Long Short-Term Memory network (sm-LSTM)
for instance-aware matching image and sentence. Our main contribution is proposing a
multimodal context-modulated attention scheme to select salient pairwise instances
from image and sentence, and a multimodal LSTM network for local similarity measurement and aggregation.
We have systematically studied the global context modulation in
the attentional procedure, and demonstrated its effectiveness with significant performance
improvement.
We have applied our model to the tasks of image annotation and retrieval,
and achieved the state-of-the-art results.
In the future, we will explore more advanced implementations of the context modulation
(in Equation \ref{eqn:e2}), and
further verify our model on more datasets.
We will also consider to jointly finetune the pretrained CNN
for better performance.


{\small
\bibliographystyle{ieee}
\bibliography{egbib}
}

\end{document}